\documentclass[11pt,a4paper]{article}

\usepackage[hyperref]{acl2019}
\usepackage[utf8]{inputenc}

\usepackage{times}
\usepackage{latexsym}
\usepackage{amsmath}
\usepackage{amsthm}
\usepackage{thmtools}
\usepackage{amsfonts}
\usepackage{amssymb}

\usepackage{url}

\usepackage{stmaryrd}
\usepackage{graphicx}

\usepackage{gb4e}
\noautomath

\aclfinalcopy % Uncomment this line for the final submission
\def\aclpaperid{1388} %  Enter the acl Paper ID here

\declaretheoremstyle[%
  spaceabove=0pt,%
  spacebelow=6pt,%
  headfont=\normalfont\itshape,%
  postheadspace=1em,%
  qed=\qedsymbol%
]{mystyle} 
\declaretheorem[name={Proof},style=mystyle,unnumbered,
]{proof}

\title{Semantic expressive capacity with bounded memory}

\author{Antoine Venant \and Alexander Koller \\
  Department of Language Science and Technology \\
  Saarland University \\
  \url{{venant|koller}@coli.uni-saarland.de}}

\date{\today}

\usepackage{enumitem}
\setlist{nolistsep,leftmargin=2em}

\usepackage{mathtools}
\usepackage{tikz}

\newcommand{\hidden}[1]{}

\newcommand{\nats}{\ensuremath{\mathbb{N}_0}}

\newcommand{\cG}{{\mathcal G}}
\newcommand{\bG}{{\mathbb G}}

\newcommand{\bT}{{\mathbf T}}

\newcommand*{\EVAL}[2][]{\llbracket#2\rrbracket\ifx\\#1\\\else_{#1}\fi}
\newcommand{\trees}[1]{\mathcal{T}_{#1}}
\newcommand{\treesv}[2]{\mathcal{T}_{#1}(#2)}
\newcommand{\contexts}[1]{\mathcal{C}_{#1}}
\newcommand*{\ALG}[1]{\mathcal{#1}}
\newcommand{\A}{\ALG{A}}

\DeclareMathOperator{\height}{\ensuremath{\mathsf{ht}}}

\newcommand{\relg}{\ensuremath{\mathcal R \mathcal E \mathcal L}}
\newcommand{\relf}{\ensuremath{\mathcal L}}
\newcommand{\rela}{\ensuremath{{\mathcal L}_{\leftrightarrow}}}

\DeclareMathOperator{\graphof}{\ensuremath{\mathsf{graph}}}

\newcommand{\satpair}[1]{\ensuremath{\mbox{BK}_{#1}}}

\newcommand{\countof}[2]{|#2|_{#1}}

\newcommand{\seps}{\ensuremath{\mathsf{Sep}}}
\DeclareMathOperator{\leftc}{\ensuremath{\mathsf{left}}}
\DeclareMathOperator{\rightc}{\ensuremath{\mathsf{right}}}

\DeclareMathOperator{\yield}{\ensuremath{\mathsf{yd}}}
\DeclareMathOperator{\cfyield}{\ensuremath{\yield_{pr}}}

\DeclareMathOperator{\htow}{\mathsf{slen}}
\newcommand{\csdoforder}[2]{\csdm{#1}^{(#2)}}

\newcommand{\lHRalg}[2]{\mathcal{H}_{#1}}
\newcommand{\HR}[1]{\mathcal{H}_{#1}}

\newcommand{\lang}{{\mathcal L}}

\newcommand{\rename}[3]{\mathsf{ren}_{\srcn{#1} \rightarrow \srcn{#2}}(#3)}

\newcommand{\forget}[2]{\mathsf{f}_{\srcn{#1}}(#2)}
\newcommand{\gmerge}{\;||\;}
\newcommand{\srcn}[1]{\ensuremath{\mathsf{#1}}}

\newcommand{\src}[1]{\ensuremath{\langle\mbox{\srcn{#1}}\rangle}}

\newcommand{\todo}[1]{\textcolor{red}{\textbf{(#1)}}}

\newcommand{\redge}[1]{\ensuremath{\xrightarrow{\mbox{\tiny #1}}}}
\newcommand{\ledge}[1]{\ensuremath{\xleftarrow{\mbox{\tiny #1}}}}

\newcommand{\selfloop}[1]{%
        \begin{tikzpicture}[inner sep=0pt, baseline=(base)]%
                \tikzstyle{place}=[circle,draw=white,fill=white,minimum size=1mm]%
   \node[place] (foo) [] {};%
   \path[->] (foo) edge  [in=0, out=135, loop right] node[above,yshift=1mm,xshift=-1mm] {{\tiny #1}} ();%
        \node (base) at (0,-.5ex) {};
        \end{tikzpicture}%
    }

\newcommand{\csd}{\ensuremath{\mathsf{CSD}}}
\newcommand{\csdl}{\langle}
\newcommand{\csdr}{\rangle}
\newcommand{\csdm}[1]{\overline{#1}}
\newcommand{\segcount}{K}

\definecolor{darkgreen}{rgb}{0,0.5,0}

\newcommand{\gtag}{\mbox{TAG}}
\newcommand{\gcftg}{\mbox{LM-CFTG}}

\newcommand{\supercsd}{\ensuremath{{\mathbf R}}}

\newtheorem{innercustomthm}{Lemma}
\newenvironment{clemma}[1]
  {\renewcommand\theinnercustomthm{#1}\innercustomthm}
  {\endinnercustomthm}

\begin{document}
\newtheorem{theorem}{Theorem}
\newtheorem{lemma}{Lemma}

\newtheorem{definition}{Definition}

\maketitle

\begin{abstract}
  We investigate the capacity of mechanisms for compositional semantic parsing to describe relations between sentences and semantic representations.
  We prove that in order to represent certain relations, mechanisms which are syntactically projective must be able to remember an unbounded number of locations in the semantic representations, where nonprojective mechanisms need not.
  This is the first result of this kind, and has consequences both for grammar-based and for neural systems.
\end{abstract}

%%% Local Variables:
%%% mode: latex
%%% TeX-master: "main"
%%% End:

\section{Introduction}

Semantic parsers which translate a sentence into a semantic
representation compositionally must recursively compute a partial
semantic representation for each node of a syntax tree. These partial
semantic representations usually contain placeholders at which
arguments and modifiers are attached in later composition
steps. Approaches to semantic parsing differ in whether they assume
that the number of placeholders is bounded or not. Lambda calculus
\cite{montague74:_englis,BlackburnB05} assumes that the number of
placeholders (lambda-bound variables) can grow unboundedly with the
length and complexity of the sentence. By contrast, many methods which are
based on unification \cite{copestake2001algebra} or graph merging
\cite{CourcelleE12,ChiangABHJK13} assume a fixed set of placeholders,
i.e.\ the number of placeholders is bounded.

Methods based on bounded placeholders are popular both in the design
of hand-written grammars \cite{Ben:Fli:Oep:02} and in semantic parsing
for graphs
\cite{PengSG15,groschwitz18:_amr_depen_parsin_typed_seman_algeb}. However,
it is not clear that all relations between language and semantic
representations can be expressed with a bounded number of
placeholders. The situation is particularly challenging when one
insists that the compositional analysis is \emph{projective} in the
sense that each composition step must combine adjacent substrings of
the input sentence. In this case, it may be impossible to combine a
semantic predicate with a distant argument immediately, forcing the
composition mechanism to use up a placeholder to remember the argument
position. If many predicates have distant arguments, this may exceed
the bounded ``memory capacity'' of the compositional mechanism.

In this paper, we show that there are relations between sentences and
semantic representations which can be described by compositional
mechanisms which are bounded and non-projective, but not by ones which
are bounded and projective. To our knowledge, this is the first result
on expressive capacity with respect to semantics -- in contrast to the
extensive literature on the expressive capacity of mechanisms which
describe just the string languages.

More precisely, we prove that
tree-adjoining grammars can describe string-graph relations using the
HR graph algebra \cite{CourcelleE12} with two sources (bounded,
non-projective) which cannot be described using linear monadic
context-free tree grammars and the HR algebra with $k$ sources, for
any fixed $k$ (bounded, projective). This result is especially
surprising because TAG and linear monadic CFTGs describe the same
string languages; thus the difference lies only in the projectivity of
the syntactic analysis.

We further prove that given certain assumptions on the alignment
between tokens in the sentence and edges in the graph, \emph{no}
generative device for projective syntax trees can simulate TAG with
two sources. This has practical consequences for the design of
transition-based semantic parsers (whether grammar-based or neural).

\textbf{Plan of the paper.} We will first explain the linguistic
background in Section~\ref{sec:background} and lay the formal
foundations in Section~\ref{sec:definitions}. We will then prove the
reduced semantic expressive capacity for aligned generative devices in
Section~\ref{sec:aligned} and for CFTGs in
Section~\ref{sec:lmcftg}. We conclude with a discussion of the
practical impact of our findings (Section~\ref{sec:conclusion}).

%%% Local Variables:
%%% mode: latex
%%% TeX-master: "main"
%%% End:

\section{Compositional semantic construction} \label{sec:background}

The Principle of Compositionality, which is widely accepted in
theoretical semantics, states that the meaning of a natural-language
expression can be determined from the meanings of its immediate
subexpressions and the way in which the subexpressions were
combined. Implementations of this principle usually assume that there
is some sort of syntax tree which describes the grammatical structure
of a sentence. A semantic representation is then calculated by
bottom-up evaluation of this syntax tree, starting with semantic
representations of the individual words and then recursively computing
a semantic representation for each node from those of its children.

\subsection{Compositional mechanisms}

Mechanisms for semantic composition will usually keep track of places
at which semantic arguments are still missing or modifiers can still
be attached. For instance, when combining the semantic representations
for ``John'' and ``sleeps'' in a derivation of ``John sleeps'', the
``subject'' argument of ``sleeps'' is filled with the meaning of
``John''. The compositional mechanism therefore assigns a semantic
representation to ``sleeps'' which has an unfilled \emph{placeholder}
for the subject.

The exact nature of the placeholder depends on the compositional
mechanism. There are two major classes in the literature.
\emph{Lambda-style} compositional mechanisms use a \emph{list} of
placeholders. For instance, lambda calculus, as used e.g.\ in Montague
Grammar \cite{montague74:_englis}, CCG
\cite{steedman01:_syntac_proces}, or linear-logic-based approaches in
LFG \cite{dalrymple95:_linear_logic_meanin_assem} might represent
``sleeps'' as $\lambda x. \mathsf{sleep}(x)$. Placeholders are
lambda-bound variables (here: $x$).

By contrast, \emph{unification-style} compositional mechanisms use
\emph{names} for placeholders. For example, a simplified form of the
Semantic Algebra used in HPSG \cite{copestake2001algebra} might
represent ``sleeps'' as the feature structure
$[\mathsf{subj}{:}\fbox{1}, \mathsf{sem}{:}
[\mathsf{pred}{:}\mathsf{sleep}, \mathsf{agent}{:}\fbox{1}]]$. This is
unified with $[\mathsf{subj}{:} \mathsf{John}]$. The placeholders are
\emph{holes} with \emph{labels} from a fixed set of argument names
(e.g.\ $\mathsf{subj}$). Named placeholders are also used in the HR
algebra \cite{CourcelleE12} and its derivatives, like Hyperedge
Replacement Grammars \cite{drewes97:_hyper,ChiangABHJK13} and the AM
algebra \cite{groschwitz18:_amr_depen_parsin_typed_seman_algeb}.

\begin{figure}
  \centering
  \includegraphics[scale=0.3]{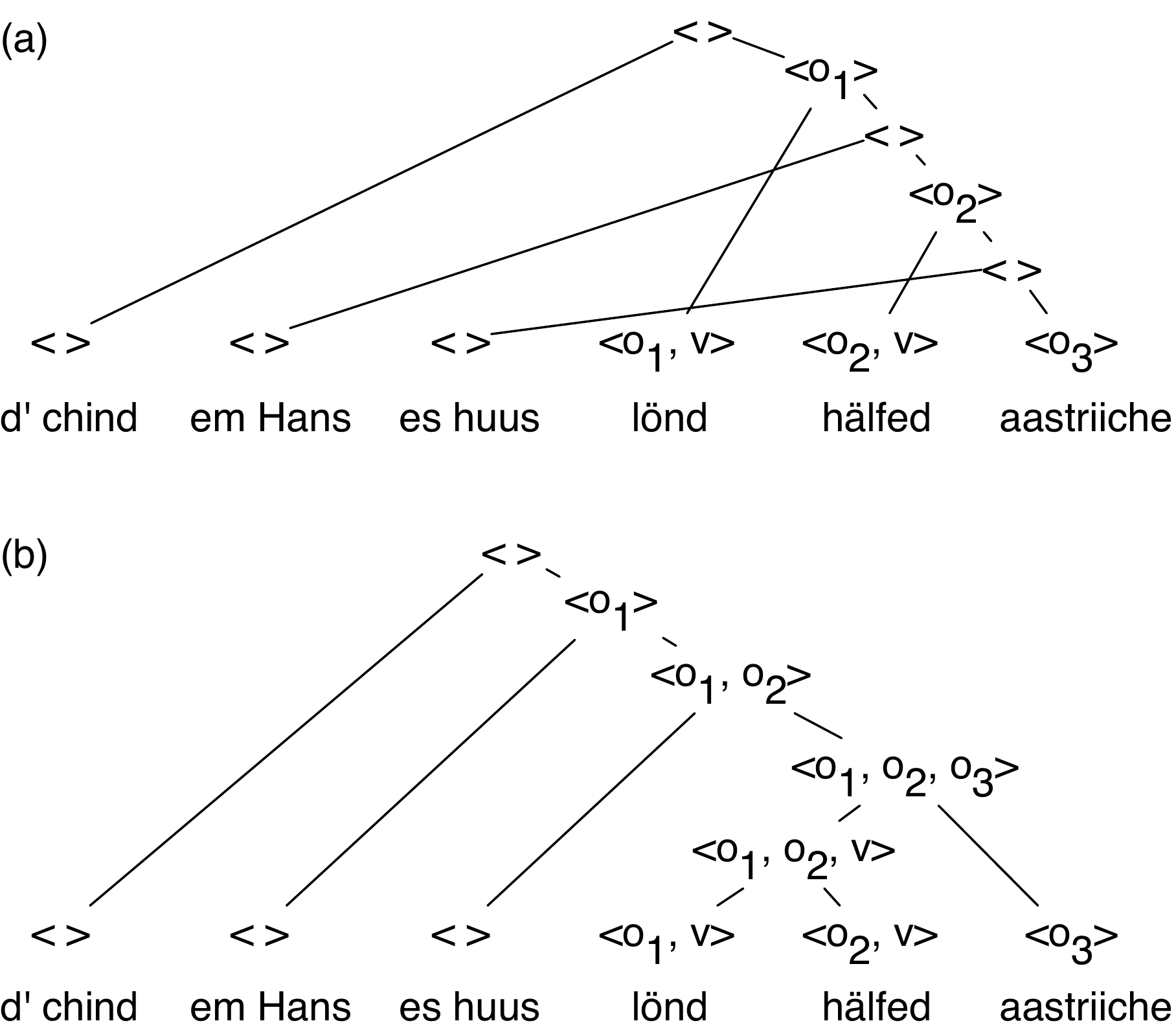}
  \strut\\[-2ex]
  \caption{(a) Nonprojective and (b) projective analysis.}
  \label{fig:projectivity}
\end{figure}

\subsection{Boundedness and projectivity}

A fundamental difference between lambda-style and unification-style
compositional mechanisms is in their ``memory capacity'': the number
of placeholders in a lambda-style mechanism can grow unboundedly with
the length and complexity of the sentence (e.g.\ by functional
composition of lambda terms), whereas in a unification-style
mechanism, the placeholders are fixed in advance.

There is an informal intuition that unbounded memory is needed
especially when an unbounded number of semantic predicates can be far
away from their arguments in the sentence, and the syntax formalism
does not allow these predicates to combine immediately with the
arguments. For illustration, consider the two derivations of the
following Swiss German sentence from \newcite{Shieber1985} in
Fig.~\ref{fig:projectivity}:

\begin{exe}
  \ex\label{ex:swiss}
  \gll (dass) (mer) {d' chind} {em Hans} {es huus} lönd hälfed aastriiche\\
       (that) (we)  {the-children-ACC} {Hans-DAT} {the-house-ACC} let help
       paint\\
  \glt `(that we) let the children help Hans paint the house'
\end{exe}

The lexical semantic representation of each verb comes with a
placeholder for its object ($o_1,o_2,o_3$) and, in the case of
``lönd'' and ``hälfed'', also one for its verb complement ($v$). The
derivation in Fig.~\ref{fig:projectivity}a immediately combines each
verb with its complements; the placeholders that are used at each node
never grow beyond the ones the verbs originally had. However, this
derivation combines verbs with nouns which are not adjacent in the
string, which is not allowed in many grammar formalisms. If we limit
ourselves to combining only adjacent substrings (\emph{projectively},
see Fig.~\ref{fig:projectivity}b), we must remember the placeholders
for all the verbs at the same time if we want to obtain the correct
predicate-argument structure. Thus, the number of placeholders grows
with the length of the sentence; this is only possible with a
lambda-style compositional mechanism.

There is scattered evidence in the literature for this tension between
bounded memory and projectivity. \newcite{ChiangABHJK13} report (of a
compositional mechanism based on the HR algebra, unification-style)
that a bounded number of placeholders suffices to derive the graphs in
the AMR version of the Geoquery corpus, but
\newcite{groschwitz18:_amr_depen_parsin_typed_seman_algeb} find that
this requires non-projective derivations in 37\% of the AMR\-Bank
training data \cite{amBanarescuBCGGHKKPS13}.
Approaches
to semantic construction with tree-adjoining grammar either perform
semantic composition along the TAG derivation tree using unification
(non-projective, unification-style)
\cite{gardent03:_seman_featur_based_tag} or along the TAG derived tree
using linear logic (projective, lambda-style) \cite{frank01:_gluet}.
\newcite{bender08:_radic_non_config_shuff_operat} discusses the
challenges involved in modeling the predicate-argument structure of a
language with very free word order (Wambaya) with projective
syntax. While the Wambaya noun phrase does not seem to require the
projective grammar to collect unbounded numbers of unfilled arguments
as in Fig.~\ref{fig:projectivity}b, Bender notes that her projective
analysis still requires a more flexible handling of semantic
arguments than the HPSG Semantic Algebra 
(unification-style) supports. 

In this paper, we define a notion of semantic expressive capacity and
prove the first formal results about the relationship between
projectivity and bounded memory.

%%% Local Variables:
%%% mode: latex
%%% TeX-master: "main"
%%% End:

\section{Formal background} \label{sec:definitions}

Let $\nats = \{0, 1, \ldots\}$ be the nonnegative integers.  A
\emph{signature} is a finite set $\Sigma$ of function symbols $f$,
each of which has been assigned a nonnegative integer called its
\emph{rank}. We write $\Sigma_n$ for the symbols of rank $n$. Given a
signature $\Sigma$, we say that all constants $a \in \Sigma_0$ are
\emph{trees} over $\Sigma$; further, if $f \in \Sigma_n$ and
$t_1,\ldots,t_n$ are trees over $\Sigma$, then $f(t_1,\ldots,t_n)$ is
also a tree. We write $\trees{\Sigma}$ for the set of all trees over
$\Sigma$. We define the \emph{height} $\height(t)$ of a tree
$t = f(t_1,\ldots,t_n)$ to be $1 + \max \height(t_i)$, and $\height(c)
= 1$ for $c \in \Sigma_0$.

Let $X \notin \Sigma$, and let $\Sigma_X = \Sigma \cup \{X\}$ (with
$X$ as a constant of rank 0). Then we call a tree
$C \in \trees{\Sigma_X}$ a \emph{context} if it contains exactly one
occurrence of $X$, and write $\contexts{\Sigma}$ for the set of all
contexts. A context can be seen as a tree with exactly one hole. If
$t \in \trees{\Sigma}$, we write $C[t]$ for the tree in
$\trees{\Sigma}$ that is obtained by replacing $X$ with $t$.

Given a string $w \in W^*$, we write $\countof{a}{w}$ for the number
of times that $a \in W$ occurs in $w$.

\begin{figure*}
  \centering
  \includegraphics[width=\textwidth]{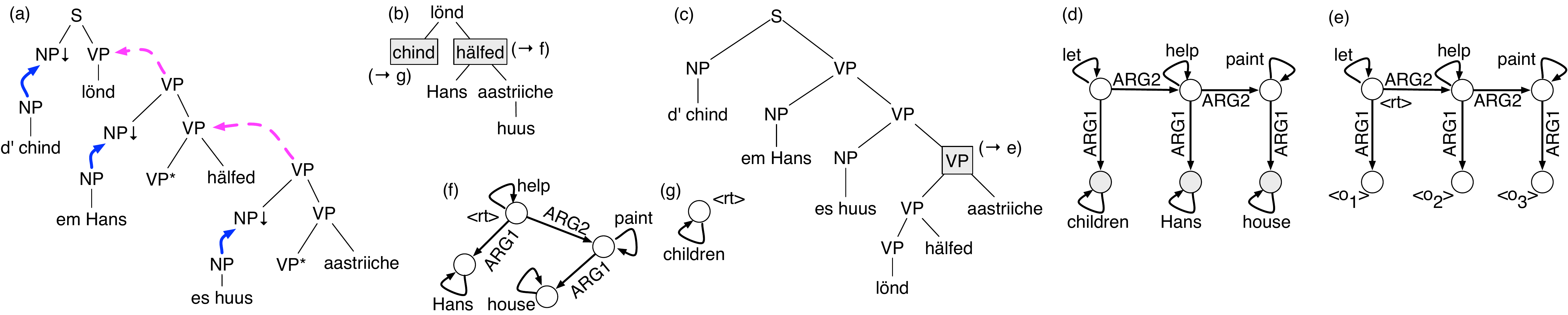}
  \strut\\[-4ex]
  \caption{Semantic construction with TAG: (a) TAG derivation, (b)
    derivation tree, (c) derived tree, (d) semantic graph.
    (e) s-graph interpretations of the
    boxed node in (c); (f,g) s-graph interpretations at the boxed
    nodes in (b).
    \label{fig:tag}}
\end{figure*}

\subsection{Grammars for strings and trees} \label{sec:grammar-formalisms}

%In this paper,
We take a very general view on how semantic
representations for strings are constructed compositionally. To this
end, we define a notion of ``grammar'' which encompasses more devices
for describing languages than just traditional grammars, such as
transition-based parsers.

We say that a \emph{tree grammar} $G$ over the signature $\Sigma$ is
any finite device that defines a language
$L(G) \subseteq \trees{\Sigma}$. For instance, regular tree grammars
\cite{ComonDGJLTL07} are tree grammars, and context-free grammars can
also be seen as tree grammars defining the language of parse trees.

We say that a \emph{string grammar} $\cG = (G,\yield)$ over the
signature $\Sigma$ and the alphabet $W$ is a pair consisting of a tree
grammar $G$ over $\Sigma$ and a \emph{yield function}
$\yield:\trees{\Sigma} \rightarrow W^*$ which maps trees to strings
over $W$ \cite{Weir88}. A string grammar defines a language
$L(\cG) = \{\yield(t) \mid t \in L(G)\} \subseteq W^*$. We call the
trees $t \in L(G)$ \emph{derivations}.

A particularly common yield function is the function $\cfyield$,
defined as
$\cfyield(f(t_1,\ldots,t_n)) = \cfyield(t_1) \cdot \ldots \cdot
\cfyield(t_n)$ if $n>0$ and $\cfyield(c) = c$ if $c$ has rank 0. This
yield function simply concatenates the words at the leaves of
$t$. Applied to the phrase-structure tree $t$ in Fig.~\ref{fig:tag}c,
$\cfyield(t)$ is the Swiss German sentence in \eqref{ex:swiss}.
Context-free grammars can be characterized as string grammars that
combine a regular tree grammar with $\cfyield$. By contrast, we can
model tree-adjoining grammars \citep[TAG,][]{joshi;etal1997} by
choosing a tree grammar $G$ that describes derivation trees as in
Fig.~\ref{fig:tag}b. The $\yield$ function could then substitute and
adjoin the elementary trees as specified by the derivation tree (see
Fig.~\ref{fig:tag}a) and then read off the words from the resulting
derived tree in Fig.~\ref{fig:tag}c.

We say that a string grammar is \emph{projective} if its yield
function is $\cfyield$. Context-free grammars as construed above are
clearly projective. Tree-adjoining grammars are \emph{not} projective:
For instance, the yield of the subtree below ``aastriiche'' in
Fig.~\ref{fig:tag}b consists of the two separate strings ``es Huus''
and ``aastriiche'', which are then wrapped around ``lönd hälfed''
further up in the derivation.

If the grammar is projective, then for any context $C$ there exist two
strings $\leftc(C)$ and $\rightc(C)$ such that for any tree $t$,
$\yield(C[t]) = \leftc(C)\cdot\yield(t)\cdot \rightc(C)$.
%We write $\leftc(C) = w_1$ and $\rightc(C) = w_2$.

\subsection{Context-free tree languages}

% We say that two string grammars $\cG, \cG'$ are \emph{weakly
%   equivalent} if $L(\cG) = L(\cG')$, i.e.\ they describe the same
% string languages. Given two sets $\bG$ and $\bG'$ of string grammars
% (for instance, two grammar formalisms), we say that $\bG$ and $\bG'$
% are weakly equivalent if
% $\{L(\cG) \mid \cG \in \bG\} = \{L(\cG') \mid \cG' \in \bG'\}$.

Below, we will talk about \emph{linear monadic context-free tree
  grammars} (\gcftg s; \newcite{rounds69:_contex},
\newcite{ComonDGJLTL07}). An \gcftg\ is a quadruple
$G = (N, \Sigma, R, S)$, where $N$ is a ranked signature of
nonterminals of rank at most one, $\Sigma$ is a ranked signature of
terminals, $S \in N_0$ is the start symbol, and $R$ is a finite set of
production rules of one of the forms
\begin{itemize}
\item $A \rightarrow t$ with $A \in N_0$ and $t \in \trees{V}$
\item $A(t) \rightarrow C[t]$ with $A \in N_1$ and $C \in \contexts{V}$,
\end{itemize}
where $V = N \cup \Sigma$. The trees in
$L(G) \subseteq \trees{\Sigma}$ are obtained by expanding $S$ with
production rules. Nonterminals of rank zero are expanded by replacing
them with trees. Nonterminals of rank one must have exactly one child
in the tree; they are replaced by a context, and the variable in the
context is replaced by the subtree below the child.  
%A tree language
%$\bT$ such that $\bT = L(G)$ for some \gcftg\ $G$ is a linear monadic
%context-free tree language (LM-CFTL).

We can extend an \gcftg\ $G$ to a string grammar $\cG =
(G,\cfyield)$. Then \gcftg\ is weakly equivalent to \gtag\
\cite{kepser11:_equiv_tree_adjoin_gramm_monad}; that is, \gcftg\ and
\gtag\ generate the same class of string languages. Intuitively, the
weakly equivalent \gcftg\ directly describes the language of derived
trees of the TAG grammar (cf.\ Fig.~\ref{fig:tag}c). Notice that
\gcftg\ is projective.

Below, we will make crucial use of the following pumping lemma for
LM-CFTLs:
\begin{lemma}[\newcite{maibaum78:_pumpin_lemmas_term_languag}]
  \label{lem:pumping}
  Let $G$ be an LM-CFTG. There exists a constant $p \in \nats$ such
  that for any $t \in L(G)$ with $\height(t) > p$, there exists a
  decomposition $t = C_1[C_2[C_3[C_4[t_5]]]]$ with
  $\height(C_2[C_3[C_4[X]]]) \leq p$ and $\height(C_2) + \height(C_4)
  > 0$ such that for any $i \in \nats$,  $C_1[v^i[t_5]] \in L(G)$,
  where we let $v^0 = C_3$ and $v^{i+1} = C_2[v^i[C_4[X]]].$

  We call $p$ the \emph{pumping height} of $L(G)$.
\end{lemma}

\subsection{The HR algebra} \label{sec:hr-algebra}

The specific unification-style semantic algebra we use in this paper
is the \emph{HR algebra} \cite{CourcelleE12}. This choice encompasses
much of the recent literature on compositional semantic parsing with
graphs, based e.g.\ on Hyperedge Replacement Grammars
\cite{ChiangABHJK13,PengSG15,Koller15} and the AM algebra
\cite{groschwitz18:_amr_depen_parsin_typed_seman_algeb}.

The values of the HR algebra are \emph{s-graphs}: directed,
edge-labeled graphs, some of whose nodes may be designated as
\emph{sources}, written in angle brackets. S-graphs can be combined
using the forget, rename, and merge operations. Rename
$\mathsf{ren}_{\srcn{a} \rightarrow \srcn{b}}$ changes an
$\srcn{a}$-source node into a $\srcn{b}$-source node. Forget
$\mathsf{f}_{\srcn{a}}$ makes it so the $\srcn{a}$-source node in the
s-graph is no longer a source node. Merge $\gmerge$ combines two
s-graphs while unifying nodes with the same source annotation. For
instance, the s-graphs $\src{rt} \redge{ARG1} \src{o}$ and
$\src{o} \selfloop{Hans}$ are merged into
$\src{rt} \redge{ARG0} \src{o} \selfloop{Hans}$.

% For instance, assume that
% $G_1 = \src{rt} \redge{ARG1} \src{o}$ and
% $G_2 = \src{rt} \selfloop{Hans}$ are two s-graphs; $G_1$ consists of
% two nodes which are a \srcn{rt}-source and an \srcn{o}-source, and
% $G_2$ consists of a single \srcn{rt}-source node with a self-loop
% edge. Then we can rename \srcn{rt} to \srcn{o} in $G_2$, obtaining
% $G_3 = \rename{rt}{o}{G_2} = \src{o} \selfloop{Hans}$. Further, the
% result of the merge operation $G_1 \gmerge G_3$ is the graph
% $G = \src{rt} \redge{ARG0} \src{o} \selfloop{Hans}$, which is obtained
% by combining $G_1$ and $G_3$ while mapping all nodes with the same
% source label to one node. The forget operation removes a source
% annotation; thus
% $\forget{o}{G} = \src{rt} \redge{ARG0} \bullet \
%selfloop{Hans}$. 
%We
%refer to \newcite{CourcelleE12} for technical details and to
%\newcite{Koller15} for details on how to use s-graphs in semantic
%parsing.

The HR algebra uses operation symbols from a ranked signature $\Delta$
to describe s-graphs syntactically. $\Delta$ contains symbols for
merge (rank 2) and the forget and rename operations (rank 1). It also
contains constants (symbols of rank 0) which denote s-graphs of the
form $\src{a} \redge{f} \src{b}$ and $\src{a} \selfloop{f}$, where
$\srcn{a},\srcn{b}$ are sources and $f$ is an edge label. Terms
$t \in \trees{\Delta}$ over this signature evaluate recursively to
s-graphs $\EVAL{t}$, as usual in an algebra. Each instance of the HR
algebra uses a fixed, finite set of $k$ source names which can be used
in the constant s-graphs and the rename and forget operations. The
class of graphs which can be expressed as values of terms over the
algebra increases with $k$. We write $\lHRalg{k}{\Delta}$ for the HR
algebra with $k$ source names (and some set of edge labels).

Let $G$ be an s-graph, and let $G'$ be a subgraph of $G$, i.e.\ a
subset of its edges. We call a node a \emph{boundary node} of $G'$ if
it is incident both to an edge in $G'$ and to an edge that is not in
$G'$. For instance, the s-graph in Fig.~\ref{fig:tag}e is a subgraph
of the one in Fig.~\ref{fig:tag}d; the boundary nodes are drawn shaded
in (d). The following lemma holds:

\begin{lemma} \label{lem:boundary} Let $G = \EVAL{C[t]}$ be an
  s-graph, and let $G'$ be a subgraph of $G$ such that the s-graph
  $\EVAL{t}$ contains the same edges as $G'$. Then every boundary node
  in $G'$ is a source in $\EVAL{t}$.
\end{lemma}

\subsection{Grammars with semantic  interpretations}
\label{sec:gramm-with-semant}

Finally, we extend string grammars to compositionally relate strings
with semantic representations. Let $\cG = (G, \yield)$ be a string
grammar. The tree grammar $G$ generates a language
$L(G) \subseteq \trees{\Sigma}$ of trees. We will map each tree
$t \in L(G)$ into a term $h(t)$ over some algebra $\A$ over a
signature $\Delta$ using a linear tree homomorphism (LTH)
$h: \trees{\Sigma} \rightarrow \trees{\Delta}$ \cite{ComonDGJLTL07},
i.e.\ by compositional bottom-up evaluation. This defines a relation
between strings and values of $\A$:
$$\relg(\cG, h, \A) = \{(\yield(t), \EVAL[\A]{h(t)}) \mid t \in
L(G)\}$$

For instance, $\A$ could be some HR algebra $\lHRalg{k}{\Delta}$; then
$\relg(\cG, h, \lHRalg{k}{\Delta})$ will be a binary relation between
strings and s-graphs. In this case, we abbreviate $\EVAL{h(t)}$ as
$\graphof(t)$.

If we look at an entire class $\bG$ of string grammars and a fixed
algebra, this defines a class of such relations:
$$\relf(\bG, \A) = \{ \relg(\cG, h, \A) \mid \cG \in \bG, \mbox{$h$
  LTH }\}.$$

% This definition is closely related to Interpreted Regular Tree
% Grammars \cite{KollerK11}, except that their definition is restricted
% to tree languages which are regular.

In the example in Fig.~\ref{fig:tag}, we can define a linear
homomorphism $h$ to map the derivation tree $t$ in (b) to a term $h(t)$
which evaluates to the s-graph shown in (d). At the top of this term,
the s-graphs at the ``chind'' and ``hälfed'' (f,g) nodes are
combined into (d) by $h(\mbox{lönd})$:
\begin{align*}
  h(\mbox{lönd}) = &\src{rt}\selfloop{let}\\[-1ex]
  & \gmerge \forget{o}{\src{rt}
  \redge{ARG1} \src{o} \gmerge \rename{rt}{o}{G_{(f)}}} \\[-1ex]
  & \gmerge \forget{o}{\src{rt}\redge{ARG2}\src{o} \gmerge \rename{rt}{o}{G_{(g)}}}
\end{align*}

This non-projective derivation produces the s-graph in (d) using only
two sources, \srcn{rt} and \srcn{o}. By contrast, a homomorphic
interpretation of the projective tree (c) has to use at least four
sources, as the intermediate result in (e) illustrates.

%%% Local Variables:
%%% mode: latex
%%% TeX-master: "main"
%%% End:

\section{Projective cross-serial dependencies} 
\label{sec:aligned}

We will now investigate the ability of projective grammar formalisms
$(\bG, \lHRalg{k}{\Delta})$ to express
$\relf(\gtag,\lHRalg{2}{\Delta})$. We will define a relation
$\csd \in \relf(\gtag,\lHRalg{2}{\Delta})$ and prove that $\csd$
cannot be generated by projective grammar formalisms with bounded
$k$. We show this first for arbitrary projective $\bG$, under certain
assumptions on the alignment of words and graph edges. In
Section~\ref{sec:lmcftg}, we drop these assumptions, but focus on
$\bG = \gcftg$.

\subsection{The relation $\csd$}
To construct $\csd$, consider the string language $\csd_s = \{A^n B^m C^n D^m \mid m,n \geq 1\}$, where 
$$A = \{ a \; \csdl^k \; \csdm{a}^k \; \csdr^k \mid k \geq 0 \},$$
and analogously for $B,C,D$. An example string in $\csd_s$ is 
$a \csdl \csdl \csdm{a} \csdm{a} \csdr \csdr \;
b \csdl \csdm{b} \csdr \;
b \;
c \csdl \csdm{c} \csdr \;
d \; d.$ 
Note that  $k$ can be chosen independently for each segment.

Every string $w \in \csd_s$ can be uniquely described by $m$, $n$, and
a sequence $\segcount(w) = (K^{(a)}, K^{(b)}, K^{(c)}, K^{(d)})$ of
numbers specifying the $k$'s used in each segment, where
$K^{(a)}, K^{(c)}$ each contain $n$ numbers and $K^{(b)}, K^{(d)}$
contain $m$ numbers. In the example, we have $n=1$, $m=2$, and
$\segcount(w) = ((2), (1, 0), (1), (0, 0))$.

\begin{figure}
  \centering
  \includegraphics[width=0.6\columnwidth]{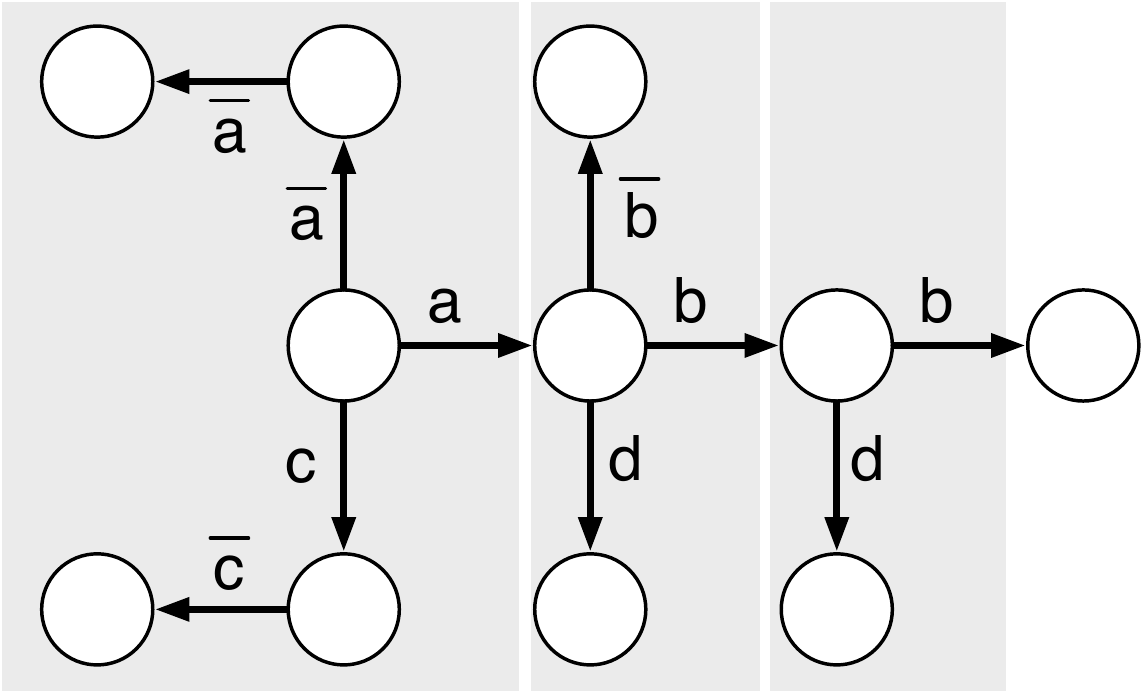}
  \caption{The $\csd$ graph for ((2), (1, 0), (1), (0, 0)); blocks
    indicated by gray boxes.}
  \label{fig:csd-graph}
\end{figure}

We associate a graph $G_w$ with each string $w \in \csd_s$ by the
construction illustrated in Fig.~\ref{fig:csd-graph}. For each
$1 \leq i \leq n$, we define the $i$-th \emph{$a$-block} to be the
graph consisting of nodes $u \redge{c} v$ with a further outgoing
$a$-edge from $u$. In addition, $u$ connects to a linear chain of
$K^{(a)}_i$ edges with label $\csdm{a}$, and $v$ to a linear chain of
$K^{(c)}_i$ $\csdm{c}$-edges. $G_w$ consists of a linear chain of the
$n$ $a$-blocks, followed by the $m$ $b$-blocks (defined
analogously). 
%The graph in Fig.~\ref{fig:csd-graph} corresponds to the
%example string $w$. 
We let $\csd = \{ (w, G_w) \mid w \in \csd_s \}$.

Note that $\csd$ is a more intricate version of the cross-serial
dependency language. $\csd$ can be generated by a TAG grammar along
the lines of the one from Section~\ref{sec:gramm-with-semant}, using a
HR algebra with two sources; thus
$\csd \in \lang(\gtag, \lHRalg{2}{\Delta})$.

%$
\subsection{\csd\ with bounded blocks}
\label{sec:boundedblocks}

The characteristic feature of \csd\ is that edges which are close
together in the graph (e.g.\ the $a$ and $c$ edge in an $a$-block)
correspond to symbols that can be distant in the string (e.g.\ $a$ and
$c$ tokens). Projective grammars cannot combine predicates
($a$) and arguments ($c$) directly because of their distance in the
string; intuitively,  they must keep track of either the $c$'s
or the $a$'s for a long time, which cannot be done with a bounded $k$.

\begin{figure}
  \centering
  \includegraphics[width=\columnwidth]{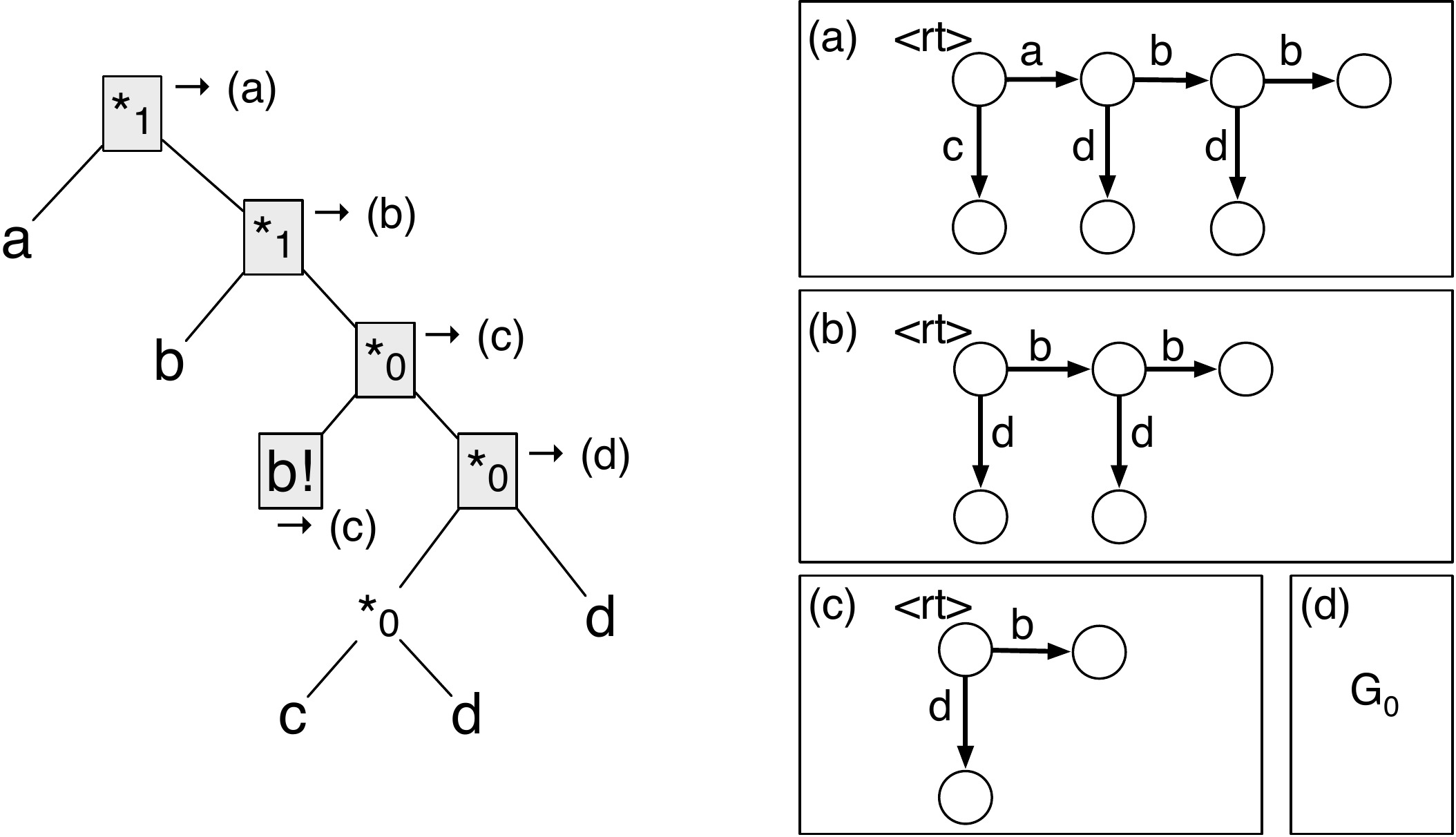}
  \caption{An  derivation of ((0), (0,0), (0), (0,0)). \label{fig:stupid}}
\end{figure}

Before we go into exploiting this intuition, we first note that its
correctness depends on the details of the construction of \csd, in
particular the ability to select arbitrary and independent $K^{(x)}$
for the different $x \in \{a,b,c,d\}$. Consider the derivation $t$ on
the left of Fig.~\ref{fig:stupid} with its projective yield $abbcdd$;
this is the case $((0), (0,0), (0), (0,0))$ of \csd, corresponding to
the \csd\ graph $G_1$ shown in Fig.~\ref{fig:stupid} (a). We can map
$t$ to this graph by applying the following linear tree homomorphism
$h$ into $\HR{2}$:\\[-1.5ex]
$$\small \arraycolsep=1.4pt
\begin{array}{rclrcl}
  h(*_1) &= &\forget{s}{x_1 \gmerge \rename{rt}{s}{x_2}} \quad\quad
  &h(*_0) &= &x_1\\
  h(b) &= &d \ledge{} b \src{rt} \redge{} \src{s}
            &h(b!) &= &d \ledge{} b \src{rt}\\
  h(a) &= &\multicolumn{4}{l}{c \ledge{} a \src{rt} \redge{} \src{s}}
\end{array}
$$
A derivation of the form $*_0(t_1,t_2)$ evaluates to the same graph as
$t_1$; the graph value of $t_2$ is ignored. Thus if we assume that the
subtree of $t$ for $cdd$ evaluates to some arbitrary graph $G_0$, the
complete derivation $t$ evaluates to $G_1$. Some intermediate results
are shown on the right of Fig.~\ref{fig:stupid}.

If we let $\csd_0$ be the subset of $\csd$ where all $K^{(x)}$ are
zero, we can generalize this construction into an LM-CFTG which
generates $\csd_0$. Thus, $\csd_0$ can be generated by a projective
grammar that is interpreted into $\HR{2}$. But note that the
derivation in Fig.~\ref{fig:stupid} is unnatural in that the symbols
in the string are not generated by the same derivation steps that
generate the graph nodes that intuitively correspond to them; for
instance, the graphs generated for the $d$ tokens are completely
irrelevant. Below, we prevent unnatural constructions like this in two
ways. We will first assume that string symbols and graph nodes must be
\emph{aligned} (Thm.~\ref{thm:aligned}). Then we will assume that the
$K^{(x)}$ can be arbitary, which allows us to drop the alignment
assumption (Thm.~\ref{thm:main}).

\subsection{$k$-distant trees}
\label{sec:unsat-deriv}

% In order to make statements about the memory capacity required to generate $\csd$, we need to bound from below the number of sources used by derivations generating the elements of $\csd$. The key ingredient for this a sufficient condition on derivations for admitting such a lower bound, which we present below.

Let $\supercsd \supseteq \csd_0$ be some relation containing at least
the string-graph pairs of $\csd_0$, e.g.\ \csd\ itself. Assume that
\supercsd\ is generated by a projective grammar $(\cG, h)$ with
$\cG = (G, \cfyield)$ and a fixed number $k$ of sources, i.e.\ we have
$\supercsd = \relg(\cG, h, \lHRalg{k}{\Delta})$.
%For the set
%$\bT = L(G)$ of derivation trees, we have
%$(\yield(t), \graphof(t)) \in \supercsd$ for all $t \in \bT$.
We will prove a contradiction.

Given a pair $(w, G_w) \in \supercsd$, we say that two edges $e,f$ in $G_w$
are \emph{equivalent}, $e \equiv f$, if they belong to the same
block. We call a derivation tree $t \in \bT = L(G)$ \emph{$k$-distant}
if $t$ has a subtree $t'$ such that we can find $k$ edges
$e_1,\ldots,e_k \in \graphof(t')$ with $e_i \not\equiv e_j$ for all
$i \neq j$ and $k$ further edges
$e'_1,\ldots,e'_k \in G_w \backslash \graphof(t')$ such that
$e_i \equiv e'_i$ for all $i$. For such trees, we have the following
 lemma.

\begin{lemma}
  A $k$-distant tree has a subtree $t'$ such that
  $\graphof(t')$  has at least $k$ sources.
   \label{lem:pairs}  
\end{lemma}
\begin{proof}
  Let $\satpair{i}$ be the $i$-th block in $G_w$; we let
  $1 \leq i \leq m+n$ and do not distinguish between $a$- and
  $b$-blocks. Let $t'$ be the subtree of $t$ claimed by the definition
  of distant trees. For each $i$, let
  $E'_i = \satpair{i} \cap \graphof(t')$ be the edges in the $i$-th
  block generated by $t'$, and let
  $E_i = \satpair{i} \backslash E'_i$.

  By definition, $E_i$ and $E'_i$ are both non-empty for at least $k$
  blocks. Each of these blocks is weakly connected,
  and thus contains at least one node $u_i$ which is incident both to
  an edge in $E_i$ and in $E'_i$. This node is a boundary node of
  $\graphof(t')$. Because $u_1,\ldots,u_k$ are all distinct, it
  follows from Lemma~\ref{lem:boundary} that $\graphof(t')$ has at
  least $k$ sources.
\end{proof}

We also note the following lemma about derivations of projective
string grammars, which follows from the inability of projective
grammars to combine distant tokens. We write
$\seps = \{a/c, c/a, b/d, d/b\}$.

\begin{lemma}
  Let $\cG = (G,\yield)$ be a projective string grammar. For any
  $r \in \nats$ there exists $s \in \nats$ such that any $t \in L(G)$
  with $\yield(t) \in a^*b^{s}c^{s}d^*$ has a subtree $t'$ such that
  $\yield(t')$ contains $r$ occurrences of $x$ and no occurrences of
  $y$, for some $x/y \in \seps$.
  \label{lem:proj}
\end{lemma}

\subsection{Projectivity and alignments}

A consequence of Lemma~\ref{lem:pairs} is that if certain string-graph
pairs in $\csd_0$\ can only be expressed with $k{+}1$-distant trees,
then $\supercsd$ (which contains these pairs as well)\ is not in
$\relf(\cG, \lHRalg{k}{\Delta})$, because $\lHRalg{k}{\Delta}$ only
admits $k$ sources.

However, as we saw in Section~\ref{sec:boundedblocks}, pairs in
$\csd_0$ can have unexpected projective derivations which make do with
a low number of sources. So let's assume for now that the string
grammar $\cG$ and the tree homomorphism $h$ produce tokens and edge
labels that fit together. Let us call $\cG,h$ \emph{aligned} if for
all constants $c \in \Sigma_0$, $\graphof(c)$ is a graph containing a
single edge with label $\yield(c)$. The derivation in
Fig.~\ref{fig:stupid} cannot be generated by an aligned grammar
because the graph for the token $b$ contains a $d$-edge. We write
$\rela(\bG,\A) = \{ \relg(\cG, h, \A) \mid \mbox{$\cG \in \bG$ and
  $\cG,h$ aligned}\}$ for the class of string-semantics relations
which can be generated with aligned grammars.

Under this assumption, it is easy to see that any relation including $\csd_0$\ (hence, \csd) cannot be
expressed with a projective grammar.

\begin{theorem} \label{thm:aligned} Let $\bG$ be any class of
  projective string grammars and $\supercsd \supseteq \csd_0$. For any
  $k$, $\supercsd \not\in \rela(\bG, \lHRalg{k}{\Delta})$.
\end{theorem}
\begin{proof}
  Assume that there is a $\cG = (G,\cfyield) \in \bG$ and an LTH $h$
  such that $\supercsd = \relg(\cG,h,\HR{k})$.  Given $k$, choose
  $s \in \nats$ such that every tree $t \in \bT = L(G)$ with
  $\yield(t) = a^s b^s c^s d^s$ has a subtree $t'$ such that
  $\yield(t')$ contains $k+1$ occurrences of $x$ and no occurrences of
  $y$, for some $x/y \in \seps$. Such an $s$ exists according to
  Lemma~\ref{lem:proj}. We can choose $t$ such that 
  $(\yield(t), \graphof(t)) \in \csd_0$.
  % Such a $t$ exists because $\csd_0 \subseteq \supercsd$. 

  Because $\cG,h$ are aligned, $\graphof(t')$ contains no $y$-edge and
  at least $k+1$ $x$-edges. Each of these $x$-edges is non-equivalent
  to all the others, and equivalent to a $y$-edge in
  $\graphof(t) \backslash \graphof(t')$, so $t$ is $k{+}1$-distant. It
  follows from Lemma~\ref{lem:pairs} that $\graphof(t')$ has $k+1$
  sources, in contradiction to the assumption that $\cG,h$ uses only
  $k$ sources.
\end{proof}

%%% Local Variables:
%%% mode: latex
%%% TeX-master: "main"
%%% End:

\section{Expressive capacity of LM-CFTG}
\label{sec:lmcftg}

Thm.~\ref{thm:aligned} is a powerful result which shows that \csd\
cannot be generated by any device for generating projective
derivations using bounded placeholder memory -- if we can assume that
tokens and edges are aligned. We will now drop this assumption and
prove that \csd\ cannot be generated using a fixed set of placeholders
using \gcftg, regardless of alignment. The basic proof idea is to
enforce a weak form of alignment through the interaction of the
pumping lemma with very long $\csdm{x}$-chains. The result is
remarkable in that \gcftg\ and \gtag\ are weakly equivalent; they only
differ in whether they must derive the strings projectively or not.

\begin{theorem}
  \label{thm:main}
  $\csd \not\in \lang(\gcftg, \lHRalg{k}{\Delta})$, for any $k$.
\end{theorem}

%The rest of this section is a proof for Thm.~\ref{thm:main}. 
% We assume that $\cG = \langle G, {\yield} \rangle$ is an \gcftg\ over
% the signature $\Sigma$ with
% $\relg(\cG, h, \lHRalg{k}{\Delta}) = \csd$, and we derive a
% contradiction. As before, we write $\bT = L(G)$.

\subsection{Asynchronous derivations}
Assume that $\csd = \relg(\cG, h, \lHRalg{k}{\Delta})$, for some $k$, with $\cG = (G, \yield)$ an LM-CFTG. Proving that this is a contradiction hinges on a somewhat technical concept of
\emph{asynchronous} derivations, which have to do with how the nodes
generating edge labels such as $\csdm{a}$ are distributed over a
derivation tree. We prove that all asynchronous derivations of certain
elements of \csd\ are distant (Lemma~\ref{lem:counts}), and that
all \gcftg\ derivations of \csd\ are asynchronous
(Lemma~\ref{lem:bingo2}), which proves
Thm.~\ref{thm:main}.

In what follows, Let $\bT = L(G)$. let us write for any tree or context $t$ and symbol $x$, $n^t_x$ as a shorthand for $\countof{x}{\yield(t)}$, $e^t_x$ for the number of $x$-edges in $\graphof(t)$ and $m^t_{\csdm{x}}$ for the maximum length of a string in $\csdm{x}^*$ which is also substring of $\yield(t)$.

\begin{definition}[$x,y,l$-asynchronous derivation]
  Let $x/y \in \seps$, $l>0$, $t \in \bT$,
  %$n_x = \countof{x}{\yield(t)}$, 
  %and $n_{\csdm{y}} = \countof{\csdm{y}}{\yield(t)}$.
  %Let $m$ be the maximum length of a string in $\csdm{x}^*$ which
  %is also substring of $\yield(t)$ and .
%  \todo{$n_{\csdm{x}}$ unused?}
  We call $t$ an \emph{$x,y,l$-asynchronous derivation} iff there is a
  decomposition $t = C[t']$ such that %$\graphof(t')$ contains at
 % most $(n_x+m)l$ $\csdm{x}$-edges and at least $(n_{\csdm{y}} - n_x
  %l)$ $\csdm{y}$-edges.
  \begin{align*}
    &e^{t'}_{\csdm{y}} \ge n^{t}_{\csdm{y}} - n^{t}_xl- m^t_{\csdm{y}}\\
    &e^{t'}_{\csdm{x}} \le  n^{t}_xl + m^t_{\csdm{x}}(l+1).
  \end{align*}
  We call the pair $(C, t')$ an \emph{$x,y,l$-asynchronous split} of $t$.
  \label{def:asyn}
\end{definition}

%Next we show that for any $l > 0$ and $k$, some elements of $\csd$ are such that $l$-asynchronicity implies $k$-unsaturation for all their derivations. 

\begin{lemma}
  For any $k,l > 0$, there is a pair
  $o_{k,l} = (w_{k,l}, G_{k,l}) \in \csd$ such that every
  $x,y,l$-asynchronous $t$ with $o_{k,l} = (\yield(t),\graphof(t))$ is
  $k$-distant. 
  %\todo{assumptions on $\yield$?} -> none.
  \label{lem:counts}
\end{lemma}
\begin{proof}
  For $x \in \{a,b,c,d\}$ and $m \in \nats$, let $\csdoforder{x}{m}$
  denote the word $\csdl^{m}\,\csdm{x}^m\,\csdr^{m}$. Let $r = s = 3l+k+1$ and
  $o_{k,l} = (w_{k,l}, G_{k,l})$ be the unique element of
  $\csd$ such that
  $$w_{k,l} = (a\csdoforder{a}{s}
  )^{r}(b\csdoforder{b}{s})^{r}(c\csdoforder{c}{s}
  )^{r}(d\csdoforder{d}{s} )^{r}.$$

  Let $t$ be an $a,c,l$-asynchronous derivation of $o_{k,l}$; other
  choices of $x/y \in \seps$ are analogous. By definition, we can
  split $t = C[t']$ such that $\graphof(t')$ has at most
  $q_a = lr+(l+1)s = (2l+1)s$
  $\csdm{a}$-edges and at least $q_c = rs-rl-s = (2l+k+1)s$
  $\csdm{c}$-edges. Notice first that $\graphof(t')$ contains at most
  $2l+1$ different complete  $a$-blocks of $G_{k,l}$,
  because each $a$-block contains $s$
  $\csdm{a}$-edges. Having $2l+2$ of them would require $(2l+2)s$ $\csdm{a}$-edges, which is more than the
  $q_a$ $\csdm{a}$-edges that $\graphof(t')$ can contain.

  Next, consider $2l + k$ distinct
  $a$-blocks of $G_{k,l}$. These blocks contain a
  total of $(2l+k)s  < (2l+k+1)s = q_c$
  $\csdm{c}$-edges. Hence, the $\csdm{c}$-edges of $\graphof(t')$
  cannot be contained within only $2l+k$ distinct blocks.

  So we
  can find at least $2l+k+1$ $\csdm{c}$-edges in $\graphof(t')$ which are
  pairwise non-equivalent. There are at least $k$ edges among these
  which are equivalent to an edge in $G_{k,l} \setminus \graphof(t')$,
  because $\graphof(t')$ contains at most $l$ complete 
  $a$-blocks of $G_{k,l}$. Thus, $t$ is $k$-distant.
 \end{proof}

%%% Local Variables:
%%% mode: latex
%%% TeX-master: "main"
%%% End:

\subsection{LM-CFTG derivations are asynchronous}

So far, we have not used the assumption that $\bT$ is an LM-CFTL. We
will now exploit the pumping lemma to
show that all derivation trees of an \gcftg\ for \csd\ must be
asynchronous.

\begin{lemma}
  If $\bT$ is an LM-CFTL, then there exists $l_0 \in \nats$ such that
  for every $t \in \bT$, there exists $x/y \in \seps$ such that $t$ is
  $x,y,l_0$-asynchronous.
  \label{lem:bingo2}
\end{lemma}

We prove this lemma by appealing to a class of derivation trees in
which predicate and argument tokens are generated in separate parts.

%We let by convention $\yield(\bot) = 0$ and for any $t_0$, $t_0[\bot]
%= t_0$. This avoids tedious case distinctions.

\begin{definition}[$x,y,l$-separated derivation]
  Let $x/y \in \seps$. A tree $t \in \trees{\Delta}$ is \emph{$x,y,l$-separated}
  if we can write $t=C_x[C_0[t_y]]$ such that
  $\countof{x}{\yield(t_y)} = 0$ and
  $\countof{y}{\yield(C_x)} = 0$ and
  $\countof{x}{\yield(C_0)} \le l$. 
  The triple $(C_x, C_0, t_y)$ is called an \emph{$l$-separation} of
  $t$.  We call an $l$-separation \emph{minimal} if there is no other
  $l$-separation of $t$ with a smaller $C_0$.
\end{definition}
%Note that any $t$ with $\countof{x}{\yield(t)} \le l$ is $l$-separated because $t = X[t[\bot]]$.  

Intuitively, we can use the pumping lemma to systematically remove
some contexts from a $t \in \bT$. From the shape of \csd, we can
conclude certain alignments between the strings and graphs generated
by these contexts and establish bounds on the number of $\csdm{x}$-
and $\csdm{y}$-edges generated by the lower part of a separated
derivation. The full proof is in the appendix; we sketch
the main ideas here.

% This allows us to prove inductively that $t$ is
%asynchronous.

Let $p$ denote the pumping height of $\bT$. There is a maximal number
of string tokens and edges that a context of height at most $p$ can
generate under a given yield and homomorphism. We call this number
$l_0$ in the rest of the proof.

\begin{lemma}
  \label{lem:xybound}
  For $t \in \bT$, let $r^t_{\csdm{y}}$ be the length of the maximal
  substring of $\yield(t)$ consisting in only $\csdm{y}$-tokens and
  containing the rightmost occurrence of $\csdm{y}$ in $\yield(t)$. If
  $t$ is $x,y, l_0$-separated, there exists a minimal
  $x,y, l_0$-separation $D_x[D_0[t_y]]$ of $t$ such that, letting
  $t_0 = D_0[t_y]$,
  $e^{t_0}_{\csdm{y}} \ge n^{t}_{\csdm{y}} - n^t_xl_0 -
  r^t_{\csdm{y}}$.

  Moreover, for any $x,y,l_0$-separation $t = E_x[E_0[t^1_y]]$, letting $t_1 = E_0[t^1_y]$, $e^{t_1}_{\csdm{x}} \le n^{t_1}_{\csdm{x}} + n^{t}_xl_0$.
\end{lemma}
\begin{proof}[sketch]
  Both statements must be achieved in separated inductions on the height of $t$, although they mostly follow similar steps. We therefore focus here only on the crucial parts of the (slightly trickier) bound on $e^{t_0}_{\csdm{y}}$. Let $D_x[D_0[t_y]]$ be a minimal $x,y,l_0$-separation of $t$ and $t_0 = D_0[t_y]$. 

  \textbf{Base Case}
  If $\height(t) \le p$, we have $n^{t}_{\csdm{y}} \le l_0$. We also have $n^t_x > 0$, so $n^{t}_{\csdm{y}} - n^t_{x}l_0 - r^t_{\csdm{y}} \le 0 \le e^{t_0}_{\bar y}$.

  \textbf{Induction step}
  If $h(t) > p$, we apply Lemma~\ref{lem:pumping} to $t$ to yield a decomposition
  $t = C_1[C_2[C_3[C_4[t_5]]]]$, where $t' = C_1[C_3[t_5]] \in \bT$,
  $\height(t') < \height(t)$ and $\height(C_2[[C_3]C_4]) \leq p$. We
  first observe that $t'$ is $x,y, l_0$-separated. By induction, there
  exists a minimal separation $t' = C_x[C_0[t'_y]]$ with $t_0' =
  C_0[t'_y]$ validating the bound on $e^{t'_0}_{\csdm{y}}$. Because of
  pumping considerations, we need to distinguish only three
  configurations of $C_2$ and $C_4$. We present only the most difficult
  case here.

  %In both cases, a minimality argument ensures that $t_0$ contains all nodes from $t'_0$, thus $e^{t_0}_{\csdm{y}} \ge e^{t'_0}_{\csdm{y}}$.
  In this case $C_2$ and $C_4$ generate only one kind of bar symbol, $\csdm{y}$, and brackets. One needs to examine all possible ways $C_2$, $C_4$ and $t_0$ may overlap. We detail the reasoning in the case where $t_0$ does not overlap with $C_2$ or $C_4$. Then, since all $y$-tokens are generated by $t_0$, projectivity of the yield and the definition of $\csd$ impose that the generated $\csdm{y}$-tokens contribute to the rightmost $y$-chain \emph{i.e.} $r^t_{\csdm{y}} = r^{t'}_{\csdm{y}} + n^{C_2[C_4]}_{\csdm{y}}$. Hence $e^{t_0}_{\csdm{y}} \ge e^{t'_0}_{\csdm{y}} \ge  n^t_{\csdm{y}} - n^{C_2[C_4]}_{\csdm{y}} + n^t_xl_0 -  r^{t_0}_{\csdm{y}} + n^{C_2[C_4]}_{\csdm{y}}$.
  %\textbf{Case 2}
  %If $C_2$ or $C_4$ generate at least some $x$ symbol, we have $n^{t}_x \ge n^{t'}_x + 1$. Finally $n^{t}_{\csdm{y}} \le n^{t'}_{\csdm{y}} + l_0$. Injecting into the bound on $e^{t_0}_{\csdm{y}}$ for $t'_0$ concludes. 
\end{proof}
  
\begin{lemma}
  For any $t \in \bT$, if $t$ is $x,y, l_0$-separated then $t$ is $x,y,l_0$-asynchronous.
\end{lemma}
\begin{proof}
  By Lemma~\ref{lem:xybound} there is a minimal $x,y,l_0$-separation
  $t=D_x[D_0[t_y]]$ such that, for $t_0 = D_0[t_y]$, the bound on
  $e^{t_0}_{\csdm{y}}$ and the bound on $e^{t_0}_{\csdm{x}}$ both
  obtain. Observe that $r^{t}_{\csdm{y}} \le m^t_{\csdm{y}}$ by
  definition, and since $t_0$ generates at most $l_0$ $x$-tokens, by
  projectivity it generates at most $(l_0+1)m^t_{\csdm{x}}$
  $\csdm{x}$-tokens (one sequence of $m^t_{\csdm{x}}$ between each
  occurrence of $x$ and the next, plus possibly one before the first
  and one after the last). Thus $t$ is $x, y, l_0$-asynchronous.
\end{proof}

\begin{lemma}
  \label{lem:sep}
  For any $t \in \bT$, $t$ is $x,y, l_0$-separated for some $x/y \in \seps$.
\end{lemma}
\begin{proof}[sketch]
  The proof proceeds by induction on the height of $t$.

  If $\height(t) \le p$, then $\countof{z}{\yield(t)} \le l_0$ for any $z \in \{a,b,c,d\}$, hence $t$ is trivially $x,y,l_0$-separated for some $x/y \in \seps$.
  
  If $h(t) > p$, Lemma~1 yields a decomposition
  $t = C_1[C_2[C_3[C_4[t_5]]]]$, where $t' = C_1[C_3[t_5]] \in \bT$,
  $\height(t') < \height(t)$ and $\height(C_2[C_3[C_4]]) \leq p$. By induction $t'$ is $x, y, l_0$-separated for some $x/y \in \seps$. Let us assume $x/y = a/c$, other cases are analoguous. The challenge is to conclude to the $x,y,l$ separation of $t$, after reinsertion of $C_2$ and $C_4$ in $t'$.

  If $C_2$ and $C_4$ generate no $a$- or $c$-token, the distribution
  of $a$- and $c$-tokens in the tree is not affected, hence $t$ is
  $a, c, l_0$-separated. Otherwise, due to pumping considerations, we
  need to distinguish three possible configurations regarding the
  shape of the yields of $C_2$ and $C_4$. We present one here, see the
  appendix for the others; they are in the same
  spirit.

  We consider the case where $\leftc(C_2)$ contains some $a$-token and no $b, c, d$-tokens, and $\yield(C_4)$ contains some $c$-token.  Assume $\leftc(C_4)$ contains some $c$. It follows that all $b$-tokens are generated by $C_3$. So $t$ has less than $l_0$ $b$-tokens, by definition of $\csd$ it has then also less than $l_0$ $d$-tokens, so $(C_1, C_2[C_3[C_4]], t_5)$ is a $d, b, l_0$-separation. Assume now that $\rightc(C_4)$ contains some $c$. It follows that $t_5$ generate no $d$-token and $C_1$ generate no $b$-token. Hence $(C_1, C_2[C_3[C_4]], t_5)$ is a $b, d, l_0$-separation.
\end{proof}

\noindent
This concludes the proof of Lemma~\ref{lem:bingo2} and Thm.~\ref{thm:main}.

\section{Conclusion} \label{sec:conclusion}

We have established a notion of expressive capacity in compositional
semantic parsing. We have proved that non-projective grammars can
express sentence-meaning relations with bounded memory that projective
ones cannot. This answers an old question in the design of compositional
systems: assuming projective syntax, lambda-style compositional
mechanisms can be more expressive than unification-style ones, which
have bounded ``memory'' for unfilled arguments.

From a theoretical perspective, the stronger result of this paper is
perhaps Thm.~\ref{thm:main}, which shows without further assumptions
that weakly equivalent grammar formalisms can differ in their semantic
expressive capacity. However, Thm.~\ref{thm:aligned} may have a
clearer practical impact on the development of compositional semantic
parsers. Consider, for instance, the case of CCG, a lexicalized
grammar formalism that has been widely used in semantic parsing
\cite{bos08:_wide_boxer,ArtziLZ15,lewis2016lstm}. While a potentially
infinite set of syntactic categories can be used in the parses of a
single CCG grammar, CCG derivations are still projective in our
sense. Thus, if one assumes that derivations should be aligned (which
is natural for a lexicalized grammar), Thm.~\ref{thm:aligned} implies
that CCG with lambda-style semantic composition is more semantically
expressive than with unification-style composition. Indeed,
lambda-style compositional mechanisms are the dominant approach in CCG
\cite{steedman01:_syntac_proces,Baldridge:2002:CCH:1073083.1073137,ArtziLZ15}.

Furthermore, under the alignment assumptions of
Section~\ref{sec:aligned}, \emph{no} unification-style compositional
mechanism can describe string-meaning relations like \csd. This
includes neural models. For instance, most transition-based parsers
\cite{Nivre08,andor16:_global_normal_trans_based_neural_networ,dyer16:_recur_neural_networ_gramm}
are projective, in that the parsing operations can only concatenate
two substrings on the top of the stack if they are adjacent in the
string. Such transition systems can therefore not be extended to
transition-based semantic parsers \cite{E17-1051} without (a) losing
expressive capacity, (b) giving up compositionality, (c) adding
mechanisms for non-projectivity \cite{gomez18}, or (d) using a
lambda-style semantic algebra. Thus our result clarifies how to build
an effective and accurate semantic parser.

We have focused on whether a grammar formalism is projective or not,
while holding the semantic algebra fixed. In the future, it would be
interesting to explore how a unification-style compositional mechanism
can be converted to a lambda-style mechanism with unbounded
placeholders. This would allow us to specify and train semantic
parsers using such abstractions, while benefiting from the efficiency
of projective parsers.

\section*{Acknowledgments} We are grateful to Emily Bender, Guy
Emerson, Meaghan Fowlie, Jonas Groschwitz, and the participants of the
DELPH-IN workshop 2018 for fruitful discussions, and to the anonymous
reviewers for their insightful feedback.

%%% Local Variables:
%%% mode: latex
%%% TeX-master: "main"
%%% End:

\bibliography{mybib}
\bibliographystyle{acl_natbib}

%\appendix

\appendix

\section{Details of the proof of Theorem 1}

\begin{clemma}{4}
  Let $\cG = (G,\yield)$ be a projective string grammar. For any
  $r \in \nats$ there exists $s \in \nats$ such that any $t \in L(G)$
  with $\yield(t) \in a^*b^{s}c^{s}d^*$ has a subtree $t'$ such that
  $\yield(t')$ contains $r$ occurrences of $x$ and no occurrences of
  $y$, for some $x/y \in \seps$.
\end{clemma}
\begin{proof}
  Depending on $\yield$, one can always choose $s > r$ such that any
  $t$ with $|\yield(t)| > 2s$ has at least one strict subtree $t'$
  with $|\yield(t')| \ge 2r$. 

  The lemma follows by induction over the height of $t$. It is
  trivially true for height 1. For the induction step, consider that
  $w' = \yield(t')$ must have at least $r$ occurrences of some letter
  because of projectivity and the shape of $\yield(t)$; assume it is
  $a$, the other cases are analogous. If $w'$ has no occurrences of
  $c$, we are done. Otherwise, by projectivity, $w'$ contains all the
  $b$'s, i.e.\ $w' \in a^* b^s c^+ d^*$. In this case, either $w'$
  contains $s>r$ occurrences of $b$ and no occurrences of $d$, in
  which case we are again done. Or it contains an occurrence of $d$;
  then $w' \in a^* b^s c^s d^*$ is in the shape required by the lemma,
  and we can apply the induction hypothesis to identify a subtree
  $t''$ of $t'$ with $r$ occurrences of some $x$ and none of the
  corresponding $y$; and $t''$ is also a subtree of $t$.
\end{proof}

%In order to apply the pumping lemma to to our problem, we need to link the height of the context of a derivation tree to the size of the string fragments it might generate:
%%I'm kipping this old version here in case we want to put an actual proof in the appendix.
\hidden{
%\begin{lemma}
%  Let $\Lambda, \Delta$ be two ranked signatures. and $h: \trees{\Lambda} \mapsto \trees{\Delta}$ be a linear homomorphism. Let $c^* = max_{\delta \in \Lambda} |\csts(h(\delta))|$ and $\rho_\Lambda$ be the maximal arity of a symbol in $\Lambda$. Finally, for $k \in \nats$, let $w_h(k) = \rho_\Lambda^k + c^*(\frac{1-\rho_\Lambda^k}{1-\rho_\Lambda})$. Then for any $u \in \treesv{\Lambda}{\vrs}$ we have \[|\csts(h(t))| \le w_h(\height(t)).\]
%  \label{lem:width}
%\end{lemma}
%}
\begin{lemma}
  To any projective yield function ${\yield}$ over some signature $\Delta$ corresponds a function ${\htow}:\nats \mapsto \nats$ with the property that for any context $u \in \treesv{\Delta}{\{x_1\}}$, if $\height(u) = q$ then $|\leftc(u)\cdot \rightc(u)| \le \htow(q)$.
  \label{lem:width}
\end{lemma}
\begin{proof}
  This is a direct consequence of the finiteness of both $\Delta$ and the strings produced by the projective yield for every symbol of $\Delta$. We refer to the appendix for a formal proof.
\end{proof}
So if a tree $t=u(t')$ is a term such that $\yield(t) = s$, then the context $u$ generates at most $\htow(u)$ letters of $s$.

}

%\begin{lemma}
%\end{lemma}

\section{Details of the proof of Theorem 2}
In all of the following, we assume that for some $k \in \nats$ we have $\csd = \relg(\cG, h, \lHRalg{k}{\Delta})$, where $\cG = (G, \yield)$ is an LM-CFTG (hence projective, \emph{i.e.} $\yield = \cfyield$). We let $\bT = L(G)$ and $p$ be the pumping height of $\bT$.

\subsection{Terminology}

Let us extend the domain of $\yield$ to contexts: for a context $C$, we let $\yield(C) = \leftc(C) \cdot \rightc(C)$. 

We say that a string $s$ is \emph{balanced} if, for any $z \in \{\csdm{a},\csdm{b}, \csdm{c}, \csdm{d}\}$ and any position $i$ in $s$ such that $s_i = z$ there are two encompassing positions $k \le i \le l$ such that $s_{[k,l]} \in \{ \csdl^n \csdm{z} \csdr^n \mid n \in \nats \}$. We say that a tree or a context is balanced if its yield is balanced. By construction, all trees of $\bT$ are balanced.

For $t \in \bT$, a \emph{pumping decomposition} of $t$ is a $5$-tuple $(C_1, C_2, C_3, C_4, t_5)$, consisting in $4$ contexts $C_1$-$C_4$ and one tree $t_5$ such that $t = C_1[C_2[C_3[C_4[t_5]]]]$,  $\height(C_2[C_3[C_4]]) \leq p$, $\height(C_2) + \height(C_4)> 0$ and for any $i \in \nats$,  $C_1[v^i[t_5]] \in \bT$,
  where we let $v^0 = C_3$ and $v^{i+1} = C_2[v^i[C_4[X]]].$

%\begin{definition}
%  We say that a subtree $t'$ of $t$ is complete (in $t$) if there is no larger subtree $t'' = C[t']$ of $t$ with $\csdl \in \leftc(C)$ or $\csdr \in \rightc(C)$ and such that $\countof{z}{\yield(C)} = 0$ for any $z \notin \{\csdl, \csdr\}$. 
%\end{definition}

\subsection{Pumping considerations}

\begin{clemma}{10}
  Let $t\in \bT$ with $\height(t) > p$, and consider a pumping decomposition $t = C_1[C_2[C_3[C_4[t_5]]]]$. Let $s = \leftc(C_2) \cdot \leftc(C_4) \cdot \rightc(C_4) \cdot \rightc(C_2) = \yield(C_2[C_4])$.  The two following propositions obtain:
  
  \begin{itemize}
  \item For any $(x,y) \in \{(a,c), (b,d)\}$, $\countof{x}{s} = \countof{y}{s}$.
  \item Let $t'=C_1[C_3[t_5]]$. For $u \in \trees{\Delta}$ and $z \in \{a,b,c,d,\csdm{a},\csdm{b}, \csdm{c}, \csdm{d}\}$ let $e^{u}_z$ denote the number of $z$-edges in $\graphof(u)$. It holds for any $z \in \{a,b,c,d,\csdm{a},\csdm{b}, \csdm{c}, \csdm{d}\}$ that $e^{t}_{z} = e^{t'}_z + \countof{z}{s}$. 
  \end{itemize}
  \label{lem:csdpump}
\end{clemma}
\begin{proof}
  Let $(x,y) \in \{(a,c), (b,d)\}$. $t \in \bT$ so $\yield(t) \in \csd_s$ which entails \begin{equation}\countof{x}{\yield(t)} = \countof{y}{\yield(t)} \label{eq:ct}.\end{equation} since $t' \in \bT$ by construction, we have $\langle \yield(t'), \graphof(t') \rangle \in \csd$. From there \begin{equation}\countof{x}{\yield(t')} = \countof{y}{\yield(t')} \label{eq:ct'}. \end{equation} But $\countof{x,y}{\yield(t)} = \countof{x,y}{\yield(t')} + \countof{x,y}{s}$. Plugging this into~\eqref{eq:ct} yields \[ \countof{x}{\yield(t')} + \countof{x}{s} = \countof{y}{\yield(t')} + \countof{y}{s}.\] Simplifying using~\eqref{eq:ct'} we find $\countof{x}{s} = \countof{y}{s}$ which establishes the first point. For the second point, we have from $\langle \yield(t), \graphof(t)\rangle \in \csd$ and by definition of $\csd$ $e^{t}_z = \countof{z}{\yield(t)} = \countof{z}{\yield(t')} + \countof{z}{s}$. Similarily since $\langle \yield(t'), \graphof(t') \rangle \in \csd$ we have $e^{t'}_z = \countof{z}{\yield(t')}$. Hence $e^t_z = e^{t'}_z + \countof{z}{s}$.
\end{proof}

We will now present a pair of lemmas stating, in formal terms, that decompositions $t = C_1[C_2[C_3[C_4[t_5]]]]$ provided by the pumping lemma all fall within a small number of configurations: 
  \begin{itemize}
  \item First, in the case where the 'pumpable' contexts $C_2$ and $C_4$ generate only `bar' tokens and brackets in $\{\csdm{a},\csdm{b},\csdm{c}, \csdm{d}, \csdl, \csdr\}^*$, we show that $\yield(C_2) \in \{\csdl, \csdr\}^*$, so that only $C_4$ is actually pumping `bar' tokens of some kind. Moreover, $t_5$ generates only 'bar' tokens and brackets as well. %and $C_3$ are both balanced and $t_5$ generates no $a$,$b$, $c$ or $d$ symbol. 
  \item Second, we explore the alternative, where the 'pumpable' contexts generate some of the `core' tokens in $\{a,b,c,d\}$, say -- for the sake of this informal presentation -- some $a$-tokens. By lemma~\ref{lem:csdpump}, they must generate as many $c$-tokens, for which we can again distinguish three possible configurations: 1.\ $a$'s and $c$'s are respectively generated on different sides of a single context ($C_2$ and/or $C_4$), but then neither $C_2$ nor $C_4$ generate any $b$ or $d$-tokens. 2.\ $C_2$ generate both $a$ and $d$-tokens (on the left and right sides respectively) and no $b$ and $c$-tokens, while $C_4$ ensures generation of corresponding $b$ and $c$-tokens (on the left and right sides respectively). 3.\ Or else, one of $C_2, C_4$ generates the $a$-tokens and no $c$,$b$ or $d$ while the other generates the corresponding $c$-tokens and no $a$, $b$ or $d$.
  \end{itemize}

  Below follows the formal presentation of these lemmas:
%\todo{pumping considerations 2}
\begin{clemma}{11}
  \label{lem:pluscase}
  Let $t\in \bT$ with $\height(t) > p$, and consider a pumping decomposition $t = C_1[C_2[C_3[C_4[t_5]]]]$ such that for all $z \in \{a,b,c,d\}$, $\countof{z}{\yield(C_2[C_4])} = 0$. There is $\csdm{x} \in \{\csdm{a}, \csdm{b}, \csdm{c}, \csdm{d}\}$ such that all of the following holds:
  \begin{enumerate}
  \item $\yield(C_2) \in \{ \csdl, \csdr\}^{*}$ and $\yield(C_4) \in \{ \csdl, \csdm{x}, \csdr \}^{*}$.
  \item Either $\leftc(C_4) \in \{\csdm{x}\}^*$ and $\countof{z}{\leftc(C_3)} = 0$ for any $z \in \{a,b,c,d\}$, or symmetrically,  $\rightc(C_4) \in \{\csdm{x}\}^*$ and $\countof{z}{\rightc(C_3)} = 0$ for any $z \in \{a,b,c,d\}$.
  \item $\countof{z}{\yield(t_5)} = 0$ for any $z \in \{a,b,c,d\}$
  \end{enumerate}
\end{clemma}
\begin{proof}
  \textbf{First point:}
  Let $s = \leftc(C_1)$ and $n_0 = \countof{\csdl}{s}$. Let $y \in \{\csdm{a},\csdm{b},\csdm{c},\csdm{d}\}$ and assume $y \in \leftc(C_2)$. Pumping $C_2$-$C_4$ $n_0 + 1$-times yields a tree $t_{n_0+1} \in \bT$ such that $s\cdot\leftc(C_2)^{n_0+1}$ is a prefix of $\yield(t_{n_0+1})$. We thus see that $t_{n_0+1}$ is not balanced, which is in contradiction with $t_{n_0+1} \in \bT$. A symmetric argument establishes that $y \notin \rightc(C_2)$.

  Assume now that there are two distinct $\csdm{x}, \csdm{y} \in \{\csdm{a},\csdm{b},\csdm{c},\csdm{d}\}$ such that $\csdm{x} \in \yield(C_4)$ and $\csdm{y} \in \yield(C_4)$. Notice that, since $C_4$ does not contain non-bar tokens, if $\csdm{x}$ and $\csdm{y}$ occur on the same side of $C_4$ (for instance $\leftc(C_4) = \langle \csdm{x} \rangle \langle \csdm{y} \rangle$) then $t \notin \bT$ because no string in $\csd_S$ admits $\leftc(C_4)$ as a substring, whereas $\yield(t)$ does. So $\csdm{x}$ and $\csdm{y}$ must occur on distinct sides. It follows that $C_4$ does not generate tokens in $\{ \csdl, \csdr \}$ either: if for instance $\leftc(C_4) = u \cdot {\csdl} \cdot \csdm{x} \cdot v$ for some strings $u$ and $v$ in $\{ \csdm{x} \csdl, \csdr \}^*$, $u\cdot {\csdl} \cdot \csdm{x}\cdot v \cdot u\cdot {\csdl} \cdot\csdm{x}\cdot v$ would be a substring of $C_1[C_2[C_2[C_3[C_4[C_4[t_5]]]]]] \in \bT$ which again is a contradiction. Let now $n_1 = \countof{\csdr}{\yield(t_5)}$. Pumping $C_2$-$C_4$ $n_1+1$ times yields a tree $t_{n_1+1} \in \bT$ with a substring of the form $\csdm{x}^{n_1 + 1}\yield(t_5) \csdm{y}^{n_1+1}$ (up to $x/y$ symmetry) which cannot be balanced, yielding a final contradiction.

  \textbf{Second point:}
  $\yield(C_4) \notin \{ \csdl, \csdr\}^{*}$, because otherwise pumping $C_2$ and $C_4$ more times than the maximum number of occurrences of a bar token in $\yield(t)$ would yield an unbalanced tree. So there is a $\csdm{x}$ such that $\csdm{x} \in \leftc(C_4)$ or $\csdm{x} \in \rightc(C_4)$. Assume for contradiction that any different token occurs on the same side of $C_4$ then $C_1[C_2[C_2[C_3[C_4[C_4[t_5]]]]]] \in \bT$ contains a substring that cannot be found in any string of $\csd$ yielding a contradiction. So $\leftc(C_4) \in \{\csdm{x}\}^*$ or $\rightc(C_4) \in \{\csdm{x}\}^*$. Assume $\leftc(C_4) \in \{\csdm{x}\}^*$, the other case is symmetric. Assume for contradiction that $\countof{z}{\leftc(C_3)} > 0$ for some $z \in \{a,b,c,d\}$. Let $n_2 = \countof{\csdl}{\yield(C_3)}$. Pumping $C_2$-$C_4$ $n_2+1$ times yields a tree $t^{n_2+1} \in \bT$ such that (by projectivity) $\yield(t^{n_2+1})$ has a substring of the form $z \cdot u \cdot \csdm{x}^{n_2+1}$ where $\countof{\csdl}{u} \le n_2$. Hence $t^{n_2+1} \in \bT$ is not balanced, yielding a contradiction. 

  \textbf{Third point:}
  Assume for contradiction that $\countof{z}{\yield(t_5)} > 0$. Assume that $\leftc(C_4) \in \{\csdm{x}\}^*$, the case $\rightc(C_4) \in \{\csdm{x}\}^*$ is symmetric, and point 2 ensures that these two cases are exhaustive. Let $n_3 = \countof{\csdr}{t_5}$ and consider the tree $t^{n_3+1} \in \bT$ obtained by pumping $C_2$-$C_4$ $n_3+1$ times. By projectivity, $\yield(t^{n_3+1})$ has a substring of the form $\csdm{x}^k \cdot u\cdot z \cdot v$ with $k \ge n_3+1$ and $\countof{\csdr}{u} \le n_3$. Hence $t^{n_3+1}$ is not balanced and $t^{n_3+1} \notin \bT$, yielding a contradiction.
  
\end{proof}

\begin{clemma}{12}
  \label{lem:corecase}
  let $t\in \bT$ with $\height(t) > p$, and consider a pumping decomposition $t = C_1[C_2[C_3[C_4[t_5]]]]$. Let $(x, y, X, Y) \in \{ (a,c,A,C), (b,d,B,D) \}$ such that $\countof{x}{\yield(C_2[C_4])} \neq 0$. One of the following obtains:
 \begin{enumerate}
 \item For some $(i,j) \in \{(2,4), (4,2) \}$, $\leftc(C_i) \in X^{+}$, $\rightc(C_i) \in Y^{+}$, $\leftc(C_j) \in X^{*}$ and $\rightc(C_j) \in Y^{*}$.
 \item $\leftc(C_2) \in A^{+}$, $\rightc(C_2) \in D^{+}$, $\leftc(C_j) \in B^{+}$ and $\rightc(C_j) \in C^{+}$.
 \item Either $\leftc(C_2) \in X^{+}$, $\rightc(C_2) = \epsilon$ and $\leftc(C_4) \cdot \rightc(C_4) \in Y^+$, or symmetrically $\leftc(C_2) = \epsilon$, $\rightc(C_2) \in Y^{+}$ and $\leftc(C_4) \cdot \rightc(C_4) \in X^+$.
 \end{enumerate}
 %Moreover, the left and right sides of $C_2[C_3[C_4[t_5]]]$ is balanced.
\end{clemma}
\begin{proof}
  All these observations follow easily from the first point of Lemma~\ref{lem:csdpump} (governing the relative number of occurrences of $a, c$-tokens on one hand and $b,d$-tokens on the other hand), projectivity, and the following observation: only one side of $C_2$ or $C_4$ cannot generate two different kinds of tokens in $\{a,b,c,d\}$ or be unbalanced. Otherwise pumping would (from projectivity) ensure that the resulting tree has a substring of a shape impossible for $\csd$ (for example, if both $a$ and $b$-tokens occur on the same side of $C_2$, pumping once produces a substring $a \cdot u \cdot b \cdot v \cdot a \cdot u \cdot b \cdot v$).

  %With a very analoguous, we observe that the left and right sides of $C_2$ and $C_4$ are balance are -- again, assuming otherwise, pumping would result in a tree with a substring incompatible with $\csd$. 
\end{proof}

\hidden{
\begin{figure*}[h!t]
  \centering
  \includegraphics[width=\textwidth]{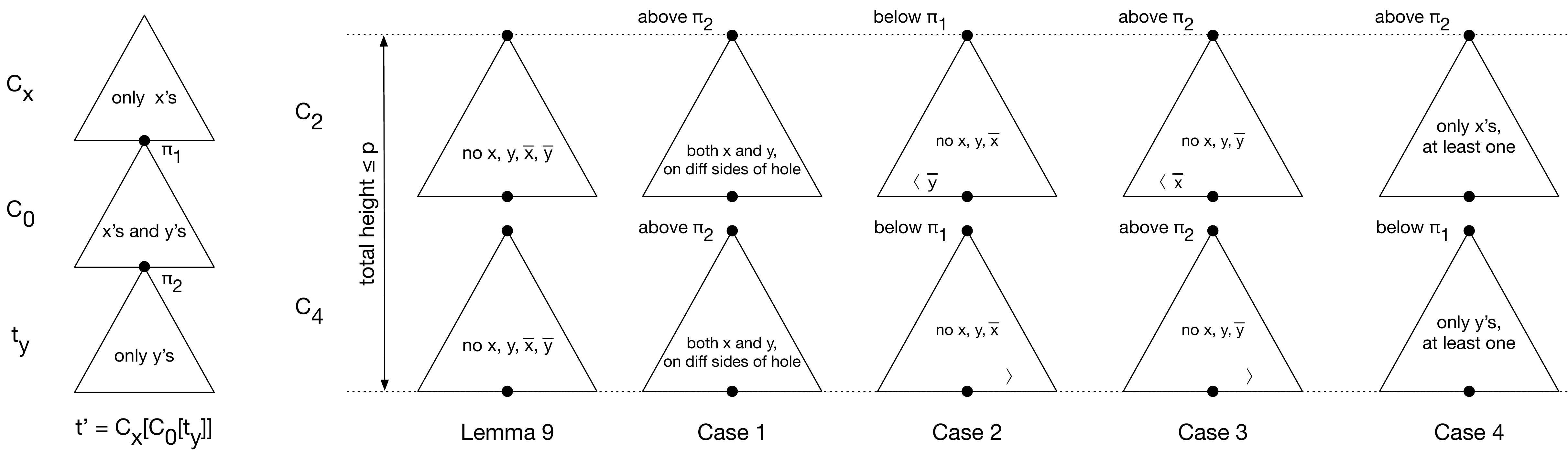}
  \caption{Possible cases for the inductive step of Lemma~\ref{lem:induction}.}
  \label{fig:cases}
\end{figure*}
}

\subsection{Separation}

\begin{clemma}{13}
  \label{lem:downwardsep}    
  Let $t = C_1[C_2[C_3[C_4[t_5]]]] \in \bT$ and $t' = C_1[C_3[t_5]] \in \bT$. If $t$ is $x, y, l$-separated then so is $t'$.
\end{clemma}
\begin{proof}
  Consider an $x,y,l$-separation of $t$: $t = D_x[D_0[t_y]]$. Let $C_x, C_0$ and $t'_y$ be respectively obtained by removing all nodes from $C_2$ or $C_4$ from $D_x$, $D_0$ and $t_y$. One easily checks that $t' = C_x[C_0[t'_y]]$.

  Moreover, $\countof{y}{\yield(C_x)} \le \countof{y}{\yield(D_x)}  = 0$, $\countof{x}{\yield(C_0)} \le \countof{x}{\yield(D_0)}  \le l$ and $\countof{x}{\yield(t'_y)} \le \countof{x}{\yield(t_y)}  = 0$. Hence $t'$ is $x, y, l$-separated.
\end{proof}

\subsection{Minimality argument}

\begin{clemma}{14}
  \label{lem:min1}
  Let $t = C_1[C_2[C_3[C_4[t_5]]]] \in \bT$ and $t' = C_1[C_3[t_5]] \in \bT$ such that $t$ is $x, y, l$-separated. By Lemma~\ref{lem:downwardsep}, $t'$ is separated. Let $D_x[D_0[t_y]]$ be a minimal separation of $t$ and $C_x[C_0[t'_y]]$ be a minimal separation of $t'$. $D_0[t_y]$ contains all nodes of $C_0[t'_y]$. 
\end{clemma}
\begin{proof}
  %\textbf{first consequent}
  Assume for contradiction that a node $\pi$ of $C_0$ is not in $D_0$. It must then be in $D_x$ or $t_y$. Assume that it is in $D_x$, the case where it is in $t_y$ is analoguous. Since $\pi$ is not in $D_0$, there is a non-trivial subcontext $D'_x$ of $D_x$ rooted at $\pi$, \emph{i.e.} $D_x = D''_x[D'_x]$ with $\height(D'_x) > 0$. Let $C''_x, C'_x$ be obtained by removing all nodes from $C_2$ or $C_4$ from $D''_x$ and $D'_x$ respectively. By definition of $D_x$, $\countof{y}{\yield(D''_x[D'_x])} = 0$, hence $\countof{y}{\yield(C''_x[C'_x])} = 0$. Further observe that we have $C_x[C_0] = C''_x[C'_x[C'_0]]$ for some subcontext $C'_0$ of $C_0$. Since $\pi$ is not in $C_2$ or $C_4$, $\height(C'_x) > 0$ thus $\height(C'_0) < \height(C_0)$. But letting $E_x = C''_x[C'_x]$, $E_x[C'_0[t'_y]]$ is then an $x, y, l$-separation of $t$ which contredicts the assumed minimality of $C_x[C_0[t'_y]]$.

  %\textbf{second consequent}
  %Assume for contradiction that a node $\pi \in C_x$ is not in $D_x$. From the previous point, $\pi$ is not a node of $C_0$ so we can write $C_x = C'_x[C_\pi]$ with $\height(C_{\pi}) > 0$. We then have $D_x = $ 
  
\end{proof}

\hidden{
\begin{clemma}{15}
  \label{lem:min2}
  Let $t = C_1[C_2[C_3[C_4[t_5]]]] \in \bT$ and $t' = C_1[C_3[t_5]] \in \bT$ such that $t$ is $x, y, l$-separated and neither $C_2$ nor $C_4$ do generate string tokens in $\{x,y\}$. By Lemma~\ref{lem:downwardsep}, $t$ is separated. Let $C_x[C_0[t'_y]]$ be a minimal separation of $t'$. There exists a minimal separation $D_x[D_0[t_y]]$ of $t$ such that $D_x$ contains all nodes of $C_x$. 
\end{clemma}
\begin{proof}
  If $\height(C_0) = 0$, let $E_x[t'_1]$ be the split of $t'$ such that the root node of $t'_1$ is the highest node of $t'_y$. One checks easily that $E_x[X[t_1]]$ is a $0$-separation of $t$ (because $C_2$ and $C_4$ do not add $x$ or $y$-tokens), minimal because $\height(X) = 0$.
  
  If $\height(C_0) \neq 0$, assume for contradiction that a node $\pi \in C_x$ is not in $D_x$ after insertion of $C_2$ and $C_4$. $\pi$ must then lie in $D_0$, for otherwise the whole subtree of $t'$ rooted at $\pi$ would generate no $x$-tokens, thus so would the subtree of $t'$ rooted at $\pi$ and, by minimality, we would have $\height(C_0) = 0$. By lemma~\ref{lem:min1}, $\pi$ is not a node of $C_0$ so we can write $C_x = C'_x[C_\pi]$ with $\height(C_{\pi}) > 0$ and $C_{\pi}$ is rooted at $\pi$. We then have $D_x[D_0] = D'_{x}[D_{\pi}[D'_0]]$ where $D'_{x}[D_{\pi}]$ is constitued of $C'_x[C_\pi]$ plus some nodes of $C_2$ and/or $C_4$. Letting $E_x = D'_{x}[D_{\pi}]$, $E_x[D'_0[t_y]]$ is an $x, y, l$-separation of $t'$ which contradicts the minimality of $D_x[D_0[t_y]]$.
\end{proof}
}

\subsection{Inductive bounds}

For any tree or context $t$ and symbol $x$, let us write $n^t_x$ as a shorthand for $\countof{x}{\yield(t)}$, $e^t_x$ for the number of $x$-edges generated by $t$ and $r^t_x$ the length of the rightmost maximal substring of $\yield(t)$ consisting in only $x$-tokens (more formally, $r^t_x = \countof{x}{s}$, where $s$ is the unique substring such that $\yield(t) = u \cdot s \cdot v$ where $s \in x^*$, if $u$ is non empty its last token is not $x$, and $\countof{x}{v} = 0$).
  
  There is a maximal number of string tokens and edges that a context of height at most $p$ can generate under the considered yield and homomorphism. We call $l_0$ this number and focus from now on $l_0$-separated and $l_0$-asynchronous derivations.

Below are the proofs of the two statements of Lemma 7 of the main paper (respectively, 7-1 and 7-2).

\begin{clemma}{7-1}
  \label{lem:xbound}
  If $t \in \bT$ is $x,y,l_0$-separated and $t = D_x[D_0[t_y]]$ is an $x,y,l_0$-separation of $t$, then for $t_0 = D_0[t_y]$ we have
  \begin{equation}
    %&e^{t_0}_{\csdm{y}} \ge n^{t}_{\csdm{y}} - n^t_xl_0 - r^t_{\csdm{y}} \tag{$y$ bound}\\
    e^{t_0}_{\csdm{x}} \le n^{t_0}_{\csdm{x}} + n^{t}_xl_0 \tag{$x$ bound}.
  \end{equation}
\end{clemma}
\begin{proof}
  We prove the result by induction over the pair $(\height(t_0), \height(t))$ (with lexicographic ordering).

  \textbf{Base Case}
  Assume $\height(t_0) \le p$. Then $e^{t_0} \le l_0$. Since $\yield(t) \in \csd_s$, $n^t_x > 0$,  thus $n^{t_0}_{\csdm{x}} + n^t_xl_0 \ge l_0$ which ensures the bound.

  \textbf{Induction step}
  If $h(t_0) > p$ then $h(t) \ge h(t_0) > p$. We apply Lemma~1 to $t$ to yield a decomposition
  $t = C_1[C_2[C_3[C_4[t_5]]]]$, where $t' = C_1[C_3[t_5]] \in \bT$,
  $\height(t') < \height(t)$ and $\height(C_2[C_3[C_4]]]) \leq p$. Notice that $t_0$ cannot overlap with $C_2[C_3[C_4]$ without overlapping with $C_1$ or $t_5$ as well, for otherwise $h(t_0) \le p$.

  As in the proof of Lemma~\ref{lem:downwardsep}, letting $C_x, C_0, t'_y$ be obtained by removing all nodes from $C_2$ and $C_4$ from $D_x$, $D_0$ and $t_y$ respectively, we obtain an $x,y,l$-separation $t' = C_x[C_0[t'_y]]$. We let $t'_0 = C_0[t'_y]$ and distinguish between possible configurations for $C_2$ and $C_4$:

  \textbf{Case 0}
  If neither $C_2$ or $C_4$ generate any $\csdm{x}$-token, we find by induction \[e^{t'_0}_{\csdm{x}} \le n^{t'_0}_{\csdm{x}} + n^{t'}_xl_0.\] Moreover, we have $e^{t'_0}_{\csdm{x}} = e^{t_0}_{\csdm{x}}$, $n^{t'_0}_{\csdm{x}} = n^{t_0}_{\csdm{x}}$ and $n^{t'}_x \le n^{t}_x$ which concludes.
  
  \textbf{Case 1}
  In this case Lemma~\ref{lem:pluscase} applies \emph{i.e.} $C_2$ and $C_4$ generate only some $\csdm{z}$-tokens and brackets. The only subcase not already covered by Case 0 is the one where $\csdm{z} = \csdm{x}$. Notice that $n^{t'}_x = n^{t}_x$. By induction, \[e^{t'_0}_{\csdm{x}} \le n^{t'_0}_{\csdm{x}} + n^{t'}_xl_0.\] If $t_0$ does not overlap with $C_2$ or $C_4$, we have $e^{t'_0}_{\csdm{x}} = e^{t_0}_{\csdm{x}}$ and $n^{t'_0}_{\csdm{x}} = n^{t_0}_{\csdm{x}}$ which ensures the bound. Otherwise $t_0$ overlaps with $C_4$. If all nodes of $t_0$ are contained in $C_4[t_5]$, then by Lemma~\ref{lem:pluscase}, $t_0$ generate no $y$-token. By separation, neither does $t$ which contradicts $t \in \csd$. Hence $t_0$ contains all nodes of $C_4$. Then by lemma~\ref{lem:pluscase} again, $n^{C_2[C_4]}_{\csdm{x}} = n^{C_4}_{\csdm{x}}$, hence $n^{t_0}_{\csdm{x}} = n^{t'_0}_{\csdm{x}} + n^{C_2[C_4]}_{\csdm{x}}$ and $e^{t_0}_{\csdm{x}} \le e^{t'_0}_{\csdm{x}} + n^{C_2[C_4]}_{\csdm{x}}$ which yields \[e^{t_0}_{\csdm{x}} \le e^{t'_0}_{\csdm{x}} + n^{C_2[C_4]}_{\csdm{x}}\le n^{t_0}_{\csdm{x}} + n^{t}_xl_0.\]

  \textbf{Case 2}
  In this case Lemma~\ref{lem:corecase} applies and at least one of  $C_2$-$C_4$ generate some token $z \in \{a,b,c,d\}$. The only subcase not already dealt with in Case 0 is the one where we can set $z = x$. We thus get inductively: \[e^{t'_0}_{\csdm{x}} \le n^{t'_0}_{\csdm{x}} + n^{t'}_xl_0.\] Since $C_2$ or $C_4$ generate at least some $x$-token, we have $n^t_x  \ge n^{t'}_x+1$. Moreover $e^{t_0}_{\csdm{x}} \le e^{t'_0}_{\csdm{x}} + l_0$ since $C_2[C_4]$ generate at most $l_0$ $\csdm{x}$-edges, and $n^{t_0}_{\csdm{x}} \ge n^{t'_0}_{\csdm{x}}$. So we have $e^{t_0}_{\csdm{x}} \le n^{t'_0}_{\csdm{x}} + n^{t'}_xl_0 + l_0 \le n^{t_0}_{\csdm{x}} + n^{t}_xl_0$ concluding the proof.  
\end{proof}

\begin{clemma}{7-2}
  \label{lem:ybound}
  If $t \in \bT$ is $x,y, l_0$-separated then $t$ there exists a minimal $x,y, l_0$-separation $D_x[D_0[t_y]]$ of $t$ is such that, letting $t_0 = D_0[t_y]$, we have
  \begin{equation}
  e^{t_0}_{\csdm{y}} \ge n^{t}_{\csdm{y}} - n^t_xl_0 - r^t_{\csdm{y}} \tag{$y$ bound}
    %&e^{t_0}_{\csdm{x}} \le n^{t_0}_{\csdm{x}} + n^{t}_xl_0 \tag{$x$ bound}.
  \end{equation}
\end{clemma}
\begin{proof}
  We prove the result by induction over the height of $t$.  

  $t$ is $x,y, l_0$-separated so let us consider $D_x[D_0[t_y]]$ a minimal $x,y, l_0$-separation of $t$. Let $t_0 = D_0[t_y]$.
  
  \textbf{Base Case}
  Assume $\height(t) \le p$. Then. $n^t_{\csdm{y}} \le l_0$. Since $\yield(t) \in \csd_s$, $n^t_x > 0$. Moreover, $0 \le e^{t_0}_{\csdm{y}}$ and $n^{t}_{\csdm{y}} \le l_0$. So $n^{t}_{\csdm{y}} - n^t_{x}l_0 - r^t_{\csdm{y}} \le 0 \le e^{t_0}_{\bar y}$ which ensures the bound.
  
  \textbf{Induction step}
  If $h(t) > p$, we apply Lemma~1 to $t$ to yield a decomposition
  $t = C_1[C_2[C_3[C_4[t_5]]]]$, where $t' = C_1[C_3[t_5]] \in \bT$,
  $\height(t') < \height(t)$ and $\height(C_2[C_3[C_4]]) \leq p$. By Lemma~\ref{lem:downwardsep}, $t'$ is $x,y, l_0$-separated. We let $t' = C_x[C_0[t'_y]]$ be a minimal separation of $t'$ verifying the bound and $t_0' = C_0[t'_y]$. In other words, we have:
  \begin{equation}
    e^{t'_0}_{\csdm{y}} \ge n^{t'}_{\csdm{y}} - n^{t'}_xl_0 - r^{t'}_{\csdm{y}} \label{eq:ybi}.
    %&e^{t'_0}_{\csdm{x}} \le n^{t'_0}_{\csdm{x}} + n^{t'}_xl_0 \label{eq:xbi}
  \end{equation}
  By Lemma~\ref{lem:min1}, $t_0 = D_0[t_y]$ contains all nodes of $t'_0$. We distinguish cases according to Lemmas~\ref{lem:pluscase}~and~\ref{lem:corecase}.
  
  \textbf{Case 1}
  Consider first the case where Lemma~\ref{lem:pluscase} applies \emph{i.e.} $C_2$ and $C_4$ generate only one kind of bar token, $\csdm{z}$, and brackets. We now distinguish cases depending on the value of $\csdm{z}$. Before this, we emphasize that in all subcases it holds that $n^{t}_{x} = n^{t'}_x$.

  \textbf{subcase i) $\csdm{z} \neq \csdm{y}$.}
  Since all nodes of $t'_0$ are contained in $t_0$, we have $e^{t_0}_{\csdm{x}} \ge e^{t'_0}_{\csdm{x}}$.  Since $C_2$ and $C_4$ generate no $\csdm{y}$-token, we have $n^{t}_{\csdm{y}} = n^{t'}_{\csdm{y}}$ and $r^t_{\csdm{y}} = r^{t'}_{\csdm{y}}$. Injecting into inequation~\eqref{eq:ybi} concludes. 
  
  \textbf{subcase ii) $\csdm{z} = \csdm{y}$.}
  We distinguish the different possible overlap of $C_2$ and $C_4$ with $t_0$.
  Notice first that, by minimality, if any $C_i$, $i \in \{2,4\}$ overlaps with $t_0$ then $t_0$ contains all nodes of $C_i$, for otherwise we would have $D_0 = D_0'[D_0'']$ with $D_0'$ a subcontext of $C_i$ such that $\height(D_0' > 0)$, and in that case $(D_x[D_0'], D_0'', t_y)$ would be a smaller $x,y,l$-separation of $t$ since $C_i$ (hence $D_0'$) does not generate $y$-tokens.
  
  Hence, in the case where $t_0$ overlaps with $C_2$, $t_0$ contains all nodes of $C_2$ and $C_4$. Since $t_0$ also contains all nodes of $t'_0$, $e^{t_0} \ge e^{t'_0}+e^{C_2[C_4]} = e^{t'_0} + n^{t}_{\csdm{y}} - n^{t'}_{\csdm{y}}$. Moreover, $r^{t}_{\csdm{y}} \ge r^{t'}_{\csdm{y}}$. We can then conclude using inequation~\ref{eq:ybi}.

  Consider now the case where $t_0$ does not overlap with $C_2$ or $C_4$. Since all $y$-tokens are generated by $t_0$, projectivity of the yield and the definition of $\csd$ impose that $r^t_{\csdm{y}} = r^{t'}_{\csdm{y}} + n^{C_2[C_4]}_{\csdm{y}}$. We further have $e^{t_0}_{\csdm{y}} \ge e^{t'_0}_{\csdm{y}}$, and injecting into inequation~\ref{eq:ybi} yields $e^{t_0}_{\csdm{y}} \ge  n^t_{\csdm{y}} - n^{C_2[C_4]}_{\csdm{y}} + n^t_xl_0 -  r^{t_0}_{\csdm{y}} + n^{[C_2[C_4]}_{\csdm{y}}$ which simplifies into the desired $y$ bound.

    Finally, in the case where $C_2$ does not overlap with $t_0$ but $C_4$ does, all nodes of $C_2$ are contained in $D_x$ and all nodes of $C_4$ are contained in $t_0$. We must then have $\countof{y}{\yield(C_3)} > 0$. Otherwise, there would exist an $x,y,l$-separation $E_x[E_0[t_y]]$ with $E_x = C_1[C_2[C_3[C_4]]]$, and $\height(E_0) < \height(D_0)$.  Assume $\countof{x}{\yield(C_3)} > 0$. Lemma~\ref{lem:pluscase}, point 2, ensures that $\countof{x,y}{\leftc(C_3)} = 0$ or $\countof{x,y}{\rightc{(C_3)}} = 0$. Assume $\countof{x,y}{\rightc(C_3)} = 0$ (the other case is symmetric). We then have both a $x$ and a $y$ generated on the left of $C_3$. Since neither $C_1[C_2]$ nor $t_5$ generate any $y$-token, projectivity imposes $r^t_{\csdm{y}} = r^{t'}_{\csdm{y}} + n^{C_2[C_4]}_{\csdm{y}}$ and we can conclude as in the previous case. The only remaining subcase is when $\countof{x}{\yield(C_3)} = 0$, in which case $t$ is $0$-separated, and considering the (minimal) $0$-separation $(C_1, X, C_2[C_3[C_4]])$ we can use the same argument as in the case where $t_0$ encompasses all nodes of $C_2$ and $C_4$.

    \textbf{Case 2}
     Consider now the remaining case where Lemma~\ref{lem:corecase} applies. If neither $C_2$ or $C_4$ generate some $x$ or $y$-token, they don't generate $\csdm{x}$ or $\csdm{y}$-tokens either, and the same reasoning as Case 1 subcase i) applies. Otherwise $C_2[C_4]$ generate at least some $x$-token. We then have $n^{t}_x \ge n^{t'}_x + 1$. Since $t_0$ contains all nodes from $t'_0$ we further have $e^{t_0}_{\csdm{y}} \ge e^{t'_0}_{\csdm{y}}$. Finally $n^{t}_{\csdm{y}} \le n^{t'}_{\csdm{y}} + l_0$. We conclude using inequation~\ref{eq:ybi}.
\end{proof}

\subsection{Conclusion}

\begin{clemma}{8}
  For any $t \in \bT$, if $t$ is $x,y, l_0$-separated then $t$ is $x,y,l_0$-asynchronous.
\end{clemma}
\begin{proof}
  By Lemma~\ref{lem:ybound}, there is a minimal $x,y,l_0$-separation $t=D_x[D_0[t_y]]$ such that the $y$ bound obtains for $t_0 = D_0[t_y]$. By lemma~\ref{lem:xbound} the $x$ bound obtains for $t_0$ as well. Observe finally, that $r^{t}_{\csdm{y}} \le m^t_{\csdm{y}}$ and since $t_0$ generates at most $l_0$ $x$-tokens, by projectivity and definition of $\csd$, it generates at most $(l_0+1)m^t_{\csdm{x}}$ $\csdm{x}$-tokens (one sequence of $m^t_{\csdm{x}}$ between each occurrence of $x$ and the next, plus possibly one in front of the first and one after the last). Hence,
  \begin{align*}
    &e^{t_0}_{\csdm{y}} \ge n^{t}_{\csdm{y}} - n^{t}_xl_0- m^t_{\csdm{y}}\\
    &e^{t_0}_{\csdm{x}} \le  n^{t}_xl + m^t_{\csdm{x}}(l_0+1).
  \end{align*}
  and $t$ is $x,y,l_0$-asynchronous.
\end{proof}

\begin{clemma}{9}
  \label{lem:sep}
  For any $t \in \bT$, $t$ is $x,y, l_0$-separated for some $x/y \in \seps$.
\end{clemma}
\begin{proof}
  The proof proceeds by induction on the height of $t$.

  If $\height(t) \le p$. Then $\countof{z}{\yield(t)} \le l_0$ for any $z \in \{a,b,c,d\}$, hence $t$ is trivially $x,y,l_0$-separated for some $x/y \in \seps$.
  
  If $h(t) > p$, Lemma~1 yields a decomposition
  $t = C_1[C_2[C_3[C_4[t_5]]]]$, where $t' = C_1[C_3[t_5]] \in \bT$,
  $\height(t') < \height(t)$ and $\height(C_2[C_4]) \leq p$. By induction $t'$ is $x,y,l_0$-separated for some $x/y \in \seps$. For sake of succintness, let us present the inductive step for $x/y = a/c$, the reasoning for other cases is analoguous. Let us examine the different possible configurations of $C_2$ and $C_4$.

  \textbf{Case 1}
  If Lemma~\ref{lem:pluscase} applies \emph{i.e.} $C_2$ and $C_4$ generate only one kind of bar token, $\csdm{z}$, and brackets, one checks easily that inserting $C_2$ and $C_4$ does not change the distribution of $a$ and $c$-tokens in the tree, hence $t$ is $a, c, l_0$-separated.

  \textbf{Case 2}
  If Lemma~\ref{lem:corecase} applies, note first that if $C_2$ and $C_4$ generate no $a$ or $c$-token, we can conclude as in Case 1 as the distribution of $a$ and $c$-tokens in the tree is not changed either. Otherwise, we assume that $C_2$ or $C_4$ generate some $a$ or $c$-token and distinguish between subcases 1-3 of Lemma~\ref{lem:corecase}: 

  \textbf{Subcase 1} in this case for some $i \in \{2,4\}$ $\leftc(C_i)$ contains an $a$-token and no $b,c$ or $d$-token while $\rightc(C_i)$ contains some $c$-token and no $a,b$ or $d$-token. Assume $i=2$, the case where $i=4$ is similar. By projectivity and definition of $\csd_s$ follows that all $b$-tokens are generated in $C_3[C_4[t_5]]$ and all $c$-tokens in $C_1$. $t$ is therefore $b,d,0$-separated, hence $b, d, l_0$-separated.
  
  \textbf{Subcase 2} in this case, $\leftc(C_2)$ contains some $a$-token and no $b, c, d$-token, $\rightc(C_3)$ contains some $d$-token and no $a, c, d$-token, $\leftc(C_4)$ contains some $b$-token and no $a, c, d$-token, $\rightc(C_4)$ contains some $c$-token and no $a, b, d$-token. It follows that $t_5$ generate no occurrence of $a$ and $C_1$ no occurrence of $c$. Since $\countof{a}{\yield(C_2[C_3[C_4]])} \le l_0$, $(C_1, C_2[C_3[C_4]], t_5)$ is an $a,c,l_0$-separation.

  \textbf{Subcase 3} Assume $\leftc(C_2)$ contains some $a$-token and no $b, c, d$-token and that $\leftc(C_4)$ contains some $c$-token. It follows that all $b$-tokens are generated by $C_3$. So $\yield(t)$ contains less than $l_0$ $b$-tokens, by definition of $\csd$ it also contains less than $l_0$ $d$-tokens, so $(C_1, C_2[C_3[C_4]], t_5)$ is a $d, b, l_0$-separation.

  Assume now $\leftc(C_2)$ contains some $a$-token and no $b, c, d$-token and that $\rightc(C_4)$ contains some $c$-token. It follows that $t_5$ generate no $d$-token and $C_1$ generate no $b$-token. Hence $(C_1, C_2[C_3[C_4]], t_5)$ is $b,d, l_0$-separation.

  The remaining cases are symmetric exchanging $c$ with $a$, $d$ with $b$, and ${\leftc}$ with ${\rightc}$ everywhere.  
\end{proof}

\end{document}

\appendix
\newtheorem{theorem}{Theorem}
\newtheorem{lemma}{Lemma}

\newtheorem{definition}{Definition}

\maketitle

\appendix

\section{Details of the proof of Theorem 1}

\begin{clemma}{4}
  Let $\cG = (G,\yield)$ be a projective string grammar. For any
  $r \in \nats$ there exists $s \in \nats$ such that any $t \in L(G)$
  with $\yield(t) \in a^*b^{s}c^{s}d^*$ has a subtree $t'$ such that
  $\yield(t')$ contains $r$ occurrences of $x$ and no occurrences of
  $y$, for some $x/y \in \seps$.
\end{clemma}
\begin{proof}
  Depending on $\yield$, one can always choose $s > r$ such that any
  $t$ with $|\yield(t)| > 2s$ has at least one strict subtree $t'$
  with $|\yield(t')| \ge 2r$. 

  The lemma follows by induction over the height of $t$. It is
  trivially true for height 1. For the induction step, consider that
  $w' = \yield(t')$ must have at least $r$ occurrences of some letter
  because of projectivity and the shape of $\yield(t)$; assume it is
  $a$, the other cases are analogous. If $w'$ has no occurrences of
  $c$, we are done. Otherwise, by projectivity, $w'$ contains all the
  $b$'s, i.e.\ $w' \in a^* b^s c^+ d^*$. In this case, either $w'$
  contains $s>r$ occurrences of $b$ and no occurrences of $d$, in
  which case we are again done. Or it contains an occurrence of $d$;
  then $w' \in a^* b^s c^s d^*$ is in the shape required by the lemma,
  and we can apply the induction hypothesis to identify a subtree
  $t''$ of $t'$ with $r$ occurrences of some $x$ and none of the
  corresponding $y$; and $t''$ is also a subtree of $t$.
\end{proof}

%In order to apply the pumping lemma to to our problem, we need to link the height of the context of a derivation tree to the size of the string fragments it might generate:
%%I'm kipping this old version here in case we want to put an actual proof in the appendix.
\hidden{
%\begin{lemma}
%  Let $\Lambda, \Delta$ be two ranked signatures. and $h: \trees{\Lambda} \mapsto \trees{\Delta}$ be a linear homomorphism. Let $c^* = max_{\delta \in \Lambda} |\csts(h(\delta))|$ and $\rho_\Lambda$ be the maximal arity of a symbol in $\Lambda$. Finally, for $k \in \nats$, let $w_h(k) = \rho_\Lambda^k + c^*(\frac{1-\rho_\Lambda^k}{1-\rho_\Lambda})$. Then for any $u \in \treesv{\Lambda}{\vrs}$ we have \[|\csts(h(t))| \le w_h(\height(t)).\]
%  \label{lem:width}
%\end{lemma}
%}
\begin{lemma}
  To any projective yield function ${\yield}$ over some signature $\Delta$ corresponds a function ${\htow}:\nats \mapsto \nats$ with the property that for any context $u \in \treesv{\Delta}{\{x_1\}}$, if $\height(u) = q$ then $|\leftc(u)\cdot \rightc(u)| \le \htow(q)$.
  \label{lem:width}
\end{lemma}
\begin{proof}
  This is a direct consequence of the finiteness of both $\Delta$ and the strings produced by the projective yield for every symbol of $\Delta$. We refer to the appendix for a formal proof.
\end{proof}
So if a tree $t=u(t')$ is a term such that $\yield(t) = s$, then the context $u$ generates at most $\htow(u)$ letters of $s$.

}

%\begin{lemma}
%\end{lemma}

\section{Details of the proof of Theorem 2}
In all of the following, we assume that for some $k \in \nats$ we have $\csd = \relg(\cG, h, \lHRalg{k}{\Delta})$, where $\cG = (G, \yield)$ is an LM-CFTG (hence projective, \emph{i.e.} $\yield = \cfyield$). We let $\bT = L(G)$ and $p$ be the pumping height of $\bT$.

\subsection{Terminology}

Let us extend the domain of $\yield$ to contexts: for a context $C$, we let $\yield(C) = \leftc(C) \cdot \rightc(C)$. 

We say that a string $s$ is \emph{balanced} if, for any $z \in \{\csdm{a},\csdm{b}, \csdm{c}, \csdm{d}\}$ and any position $i$ in $s$ such that $s_i = z$ there are two encompassing positions $k \le i \le l$ such that $s_{[k,l]} \in \{ \csdl^n \csdm{z} \csdr^n \mid n \in \nats \}$. We say that a tree or a context is balanced if its yield is balanced. By construction, all trees of $\bT$ are balanced.

For $t \in \bT$, a \emph{pumping decomposition} of $t$ is a $5$-tuple $(C_1, C_2, C_3, C_4, t_5)$, consisting in $4$ contexts $C_1$-$C_4$ and one tree $t_5$ such that $t = C_1[C_2[C_3[C_4[t_5]]]]$,  $\height(C_2[C_3[C_4]]) \leq p$, $\height(C_2) + \height(C_4)> 0$ and for any $i \in \nats$,  $C_1[v^i[t_5]] \in \bT$,
  where we let $v^0 = C_3$ and $v^{i+1} = C_2[v^i[C_4[X]]].$

%\begin{definition}
%  We say that a subtree $t'$ of $t$ is complete (in $t$) if there is no larger subtree $t'' = C[t']$ of $t$ with $\csdl \in \leftc(C)$ or $\csdr \in \rightc(C)$ and such that $\countof{z}{\yield(C)} = 0$ for any $z \notin \{\csdl, \csdr\}$. 
%\end{definition}

\subsection{Pumping considerations}

\begin{clemma}{10}
  Let $t\in \bT$ with $\height(t) > p$, and consider a pumping decomposition $t = C_1[C_2[C_3[C_4[t_5]]]]$. Let $s = \leftc(C_2) \cdot \leftc(C_4) \cdot \rightc(C_4) \cdot \rightc(C_2) = \yield(C_2[C_4])$.  The two following propositions obtain:
  
  \begin{itemize}
  \item For any $(x,y) \in \{(a,c), (b,d)\}$, $\countof{x}{s} = \countof{y}{s}$.
  \item Let $t'=C_1[C_3[t_5]]$. For $u \in \trees{\Delta}$ and $z \in \{a,b,c,d,\csdm{a},\csdm{b}, \csdm{c}, \csdm{d}\}$ let $e^{u}_z$ denote the number of $z$-edges in $\graphof(u)$. It holds for any $z \in \{a,b,c,d,\csdm{a},\csdm{b}, \csdm{c}, \csdm{d}\}$ that $e^{t}_{z} = e^{t'}_z + \countof{z}{s}$. 
  \end{itemize}
  \label{lem:csdpump}
\end{clemma}
\begin{proof}
  Let $(x,y) \in \{(a,c), (b,d)\}$. $t \in \bT$ so $\yield(t) \in \csd_s$ which entails \begin{equation}\countof{x}{\yield(t)} = \countof{y}{\yield(t)} \label{eq:ct}.\end{equation} since $t' \in \bT$ by construction, we have $\langle \yield(t'), \graphof(t') \rangle \in \csd$. From there \begin{equation}\countof{x}{\yield(t')} = \countof{y}{\yield(t')} \label{eq:ct'}. \end{equation} But $\countof{x,y}{\yield(t)} = \countof{x,y}{\yield(t')} + \countof{x,y}{s}$. Plugging this into~\eqref{eq:ct} yields \[ \countof{x}{\yield(t')} + \countof{x}{s} = \countof{y}{\yield(t')} + \countof{y}{s}.\] Simplifying using~\eqref{eq:ct'} we find $\countof{x}{s} = \countof{y}{s}$ which establishes the first point. For the second point, we have from $\langle \yield(t), \graphof(t)\rangle \in \csd$ and by definition of $\csd$ $e^{t}_z = \countof{z}{\yield(t)} = \countof{z}{\yield(t')} + \countof{z}{s}$. Similarily since $\langle \yield(t'), \graphof(t') \rangle \in \csd$ we have $e^{t'}_z = \countof{z}{\yield(t')}$. Hence $e^t_z = e^{t'}_z + \countof{z}{s}$.
\end{proof}

We will now present a pair of lemmas stating, in formal terms, that decompositions $t = C_1[C_2[C_3[C_4[t_5]]]]$ provided by the pumping lemma all fall within a small number of configurations: 
  \begin{itemize}
  \item First, in the case where the 'pumpable' contexts $C_2$ and $C_4$ generate only `bar' tokens and brackets in $\{\csdm{a},\csdm{b},\csdm{c}, \csdm{d}, \csdl, \csdr\}^*$, we show that $\yield(C_2) \in \{\csdl, \csdr\}^*$, so that only $C_4$ is actually pumping `bar' tokens of some kind. Moreover, $t_5$ generates only 'bar' tokens and brackets as well. %and $C_3$ are both balanced and $t_5$ generates no $a$,$b$, $c$ or $d$ symbol. 
  \item Second, we explore the alternative, where the 'pumpable' contexts generate some of the `core' tokens in $\{a,b,c,d\}$, say -- for the sake of this informal presentation -- some $a$-tokens. By lemma~\ref{lem:csdpump}, they must generate as many $c$-tokens, for which we can again distinguish three possible configurations: 1.\ $a$'s and $c$'s are respectively generated on different sides of a single context ($C_2$ and/or $C_4$), but then neither $C_2$ nor $C_4$ generate any $b$ or $d$-tokens. 2.\ $C_2$ generate both $a$ and $d$-tokens (on the left and right sides respectively) and no $b$ and $c$-tokens, while $C_4$ ensures generation of corresponding $b$ and $c$-tokens (on the left and right sides respectively). 3.\ Or else, one of $C_2, C_4$ generates the $a$-tokens and no $c$,$b$ or $d$ while the other generates the corresponding $c$-tokens and no $a$, $b$ or $d$.
  \end{itemize}

  Below follows the formal presentation of these lemmas:
%\todo{pumping considerations 2}
\begin{clemma}{11}
  \label{lem:pluscase}
  Let $t\in \bT$ with $\height(t) > p$, and consider a pumping decomposition $t = C_1[C_2[C_3[C_4[t_5]]]]$ such that for all $z \in \{a,b,c,d\}$, $\countof{z}{\yield(C_2[C_4])} = 0$. There is $\csdm{x} \in \{\csdm{a}, \csdm{b}, \csdm{c}, \csdm{d}\}$ such that all of the following holds:
  \begin{enumerate}
  \item $\yield(C_2) \in \{ \csdl, \csdr\}^{*}$ and $\yield(C_4) \in \{ \csdl, \csdm{x}, \csdr \}^{*}$.
  \item Either $\leftc(C_4) \in \{\csdm{x}\}^*$ and $\countof{z}{\leftc(C_3)} = 0$ for any $z \in \{a,b,c,d\}$, or symmetrically,  $\rightc(C_4) \in \{\csdm{x}\}^*$ and $\countof{z}{\rightc(C_3)} = 0$ for any $z \in \{a,b,c,d\}$.
  \item $\countof{z}{\yield(t_5)} = 0$ for any $z \in \{a,b,c,d\}$
  \end{enumerate}
\end{clemma}
\begin{proof}
  \textbf{First point:}
  Let $s = \leftc(C_1)$ and $n_0 = \countof{\csdl}{s}$. Let $y \in \{\csdm{a},\csdm{b},\csdm{c},\csdm{d}\}$ and assume $y \in \leftc(C_2)$. Pumping $C_2$-$C_4$ $n_0 + 1$-times yields a tree $t_{n_0+1} \in \bT$ such that $s\cdot\leftc(C_2)^{n_0+1}$ is a prefix of $\yield(t_{n_0+1})$. We thus see that $t_{n_0+1}$ is not balanced, which is in contradiction with $t_{n_0+1} \in \bT$. A symmetric argument establishes that $y \notin \rightc(C_2)$.

  Assume now that there are two distinct $\csdm{x}, \csdm{y} \in \{\csdm{a},\csdm{b},\csdm{c},\csdm{d}\}$ such that $\csdm{x} \in \yield(C_4)$ and $\csdm{y} \in \yield(C_4)$. Notice that, since $C_4$ does not contain non-bar tokens, if $\csdm{x}$ and $\csdm{y}$ occur on the same side of $C_4$ (for instance $\leftc(C_4) = \langle \csdm{x} \rangle \langle \csdm{y} \rangle$) then $t \notin \bT$ because no string in $\csd_S$ admits $\leftc(C_4)$ as a substring, whereas $\yield(t)$ does. So $\csdm{x}$ and $\csdm{y}$ must occur on distinct sides. It follows that $C_4$ does not generate tokens in $\{ \csdl, \csdr \}$ either: if for instance $\leftc(C_4) = u \cdot {\csdl} \cdot \csdm{x} \cdot v$ for some strings $u$ and $v$ in $\{ \csdm{x} \csdl, \csdr \}^*$, $u\cdot {\csdl} \cdot \csdm{x}\cdot v \cdot u\cdot {\csdl} \cdot\csdm{x}\cdot v$ would be a substring of $C_1[C_2[C_2[C_3[C_4[C_4[t_5]]]]]] \in \bT$ which again is a contradiction. Let now $n_1 = \countof{\csdr}{\yield(t_5)}$. Pumping $C_2$-$C_4$ $n_1+1$ times yields a tree $t_{n_1+1} \in \bT$ with a substring of the form $\csdm{x}^{n_1 + 1}\yield(t_5) \csdm{y}^{n_1+1}$ (up to $x/y$ symmetry) which cannot be balanced, yielding a final contradiction.

  \textbf{Second point:}
  $\yield(C_4) \notin \{ \csdl, \csdr\}^{*}$, because otherwise pumping $C_2$ and $C_4$ more times than the maximum number of occurrences of a bar token in $\yield(t)$ would yield an unbalanced tree. So there is a $\csdm{x}$ such that $\csdm{x} \in \leftc(C_4)$ or $\csdm{x} \in \rightc(C_4)$. Assume for contradiction that any different token occurs on the same side of $C_4$ then $C_1[C_2[C_2[C_3[C_4[C_4[t_5]]]]]] \in \bT$ contains a substring that cannot be found in any string of $\csd$ yielding a contradiction. So $\leftc(C_4) \in \{\csdm{x}\}^*$ or $\rightc(C_4) \in \{\csdm{x}\}^*$. Assume $\leftc(C_4) \in \{\csdm{x}\}^*$, the other case is symmetric. Assume for contradiction that $\countof{z}{\leftc(C_3)} > 0$ for some $z \in \{a,b,c,d\}$. Let $n_2 = \countof{\csdl}{\yield(C_3)}$. Pumping $C_2$-$C_4$ $n_2+1$ times yields a tree $t^{n_2+1} \in \bT$ such that (by projectivity) $\yield(t^{n_2+1})$ has a substring of the form $z \cdot u \cdot \csdm{x}^{n_2+1}$ where $\countof{\csdl}{u} \le n_2$. Hence $t^{n_2+1} \in \bT$ is not balanced, yielding a contradiction. 

  \textbf{Third point:}
  Assume for contradiction that $\countof{z}{\yield(t_5)} > 0$. Assume that $\leftc(C_4) \in \{\csdm{x}\}^*$, the case $\rightc(C_4) \in \{\csdm{x}\}^*$ is symmetric, and point 2 ensures that these two cases are exhaustive. Let $n_3 = \countof{\csdr}{t_5}$ and consider the tree $t^{n_3+1} \in \bT$ obtained by pumping $C_2$-$C_4$ $n_3+1$ times. By projectivity, $\yield(t^{n_3+1})$ has a substring of the form $\csdm{x}^k \cdot u\cdot z \cdot v$ with $k \ge n_3+1$ and $\countof{\csdr}{u} \le n_3$. Hence $t^{n_3+1}$ is not balanced and $t^{n_3+1} \notin \bT$, yielding a contradiction.
  
\end{proof}

\begin{clemma}{12}
  \label{lem:corecase}
  let $t\in \bT$ with $\height(t) > p$, and consider a pumping decomposition $t = C_1[C_2[C_3[C_4[t_5]]]]$. Let $(x, y, X, Y) \in \{ (a,c,A,C), (b,d,B,D) \}$ such that $\countof{x}{\yield(C_2[C_4])} \neq 0$. One of the following obtains:
 \begin{enumerate}
 \item For some $(i,j) \in \{(2,4), (4,2) \}$, $\leftc(C_i) \in X^{+}$, $\rightc(C_i) \in Y^{+}$, $\leftc(C_j) \in X^{*}$ and $\rightc(C_j) \in Y^{*}$.
 \item $\leftc(C_2) \in A^{+}$, $\rightc(C_2) \in D^{+}$, $\leftc(C_j) \in B^{+}$ and $\rightc(C_j) \in C^{+}$.
 \item Either $\leftc(C_2) \in X^{+}$, $\rightc(C_2) = \epsilon$ and $\leftc(C_4) \cdot \rightc(C_4) \in Y^+$, or symmetrically $\leftc(C_2) = \epsilon$, $\rightc(C_2) \in Y^{+}$ and $\leftc(C_4) \cdot \rightc(C_4) \in X^+$.
 \end{enumerate}
 %Moreover, the left and right sides of $C_2[C_3[C_4[t_5]]]$ is balanced.
\end{clemma}
\begin{proof}
  All these observations follow easily from the first point of Lemma~\ref{lem:csdpump} (governing the relative number of occurrences of $a, c$-tokens on one hand and $b,d$-tokens on the other hand), projectivity, and the following observation: only one side of $C_2$ or $C_4$ cannot generate two different kinds of tokens in $\{a,b,c,d\}$ or be unbalanced. Otherwise pumping would (from projectivity) ensure that the resulting tree has a substring of a shape impossible for $\csd$ (for example, if both $a$ and $b$-tokens occur on the same side of $C_2$, pumping once produces a substring $a \cdot u \cdot b \cdot v \cdot a \cdot u \cdot b \cdot v$).

  %With a very analoguous, we observe that the left and right sides of $C_2$ and $C_4$ are balance are -- again, assuming otherwise, pumping would result in a tree with a substring incompatible with $\csd$. 
\end{proof}

\hidden{
\begin{figure*}[h!t]
  \centering
  \includegraphics[width=\textwidth]{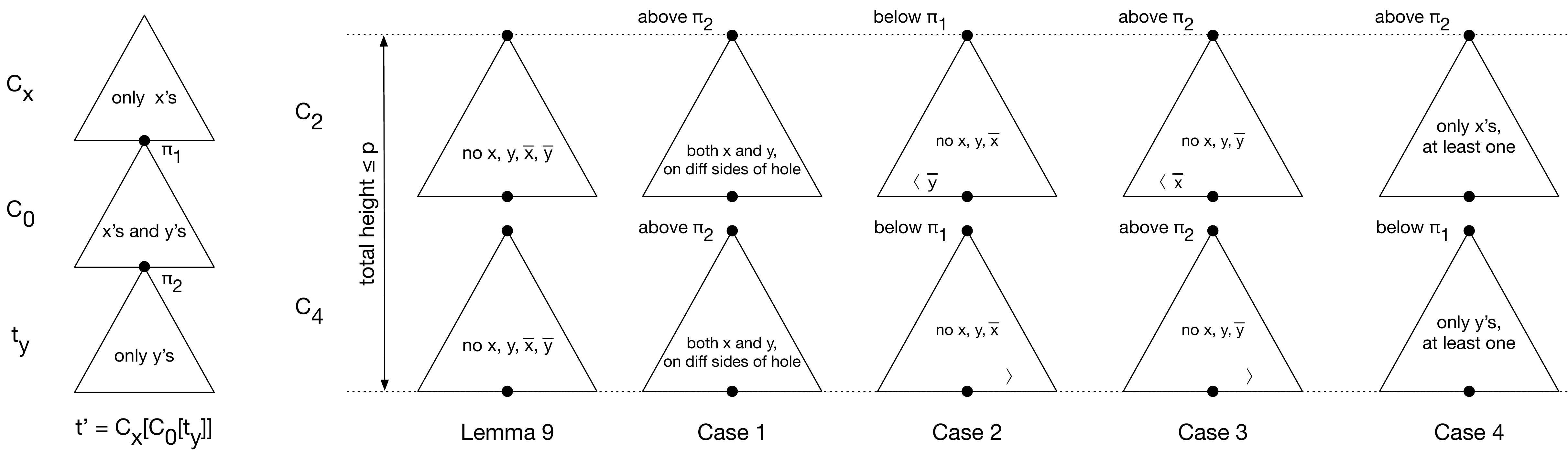}
  \caption{Possible cases for the inductive step of Lemma~\ref{lem:induction}.}
  \label{fig:cases}
\end{figure*}
}

\subsection{Separation}

\begin{clemma}{13}
  \label{lem:downwardsep}    
  Let $t = C_1[C_2[C_3[C_4[t_5]]]] \in \bT$ and $t' = C_1[C_3[t_5]] \in \bT$. If $t$ is $x, y, l$-separated then so is $t'$.
\end{clemma}
\begin{proof}
  Consider an $x,y,l$-separation of $t$: $t = D_x[D_0[t_y]]$. Let $C_x, C_0$ and $t'_y$ be respectively obtained by removing all nodes from $C_2$ or $C_4$ from $D_x$, $D_0$ and $t_y$. One easily checks that $t' = C_x[C_0[t'_y]]$.

  Moreover, $\countof{y}{\yield(C_x)} \le \countof{y}{\yield(D_x)}  = 0$, $\countof{x}{\yield(C_0)} \le \countof{x}{\yield(D_0)}  \le l$ and $\countof{x}{\yield(t'_y)} \le \countof{x}{\yield(t_y)}  = 0$. Hence $t'$ is $x, y, l$-separated.
\end{proof}

\subsection{Minimality argument}

\begin{clemma}{14}
  \label{lem:min1}
  Let $t = C_1[C_2[C_3[C_4[t_5]]]] \in \bT$ and $t' = C_1[C_3[t_5]] \in \bT$ such that $t$ is $x, y, l$-separated. By Lemma~\ref{lem:downwardsep}, $t'$ is separated. Let $D_x[D_0[t_y]]$ be a minimal separation of $t$ and $C_x[C_0[t'_y]]$ be a minimal separation of $t'$. $D_0[t_y]$ contains all nodes of $C_0[t'_y]$. 
\end{clemma}
\begin{proof}
  %\textbf{first consequent}
  Assume for contradiction that a node $\pi$ of $C_0$ is not in $D_0$. It must then be in $D_x$ or $t_y$. Assume that it is in $D_x$, the case where it is in $t_y$ is analoguous. Since $\pi$ is not in $D_0$, there is a non-trivial subcontext $D'_x$ of $D_x$ rooted at $\pi$, \emph{i.e.} $D_x = D''_x[D'_x]$ with $\height(D'_x) > 0$. Let $C''_x, C'_x$ be obtained by removing all nodes from $C_2$ or $C_4$ from $D''_x$ and $D'_x$ respectively. By definition of $D_x$, $\countof{y}{\yield(D''_x[D'_x])} = 0$, hence $\countof{y}{\yield(C''_x[C'_x])} = 0$. Further observe that we have $C_x[C_0] = C''_x[C'_x[C'_0]]$ for some subcontext $C'_0$ of $C_0$. Since $\pi$ is not in $C_2$ or $C_4$, $\height(C'_x) > 0$ thus $\height(C'_0) < \height(C_0)$. But letting $E_x = C''_x[C'_x]$, $E_x[C'_0[t'_y]]$ is then an $x, y, l$-separation of $t$ which contredicts the assumed minimality of $C_x[C_0[t'_y]]$.

  %\textbf{second consequent}
  %Assume for contradiction that a node $\pi \in C_x$ is not in $D_x$. From the previous point, $\pi$ is not a node of $C_0$ so we can write $C_x = C'_x[C_\pi]$ with $\height(C_{\pi}) > 0$. We then have $D_x = $ 
  
\end{proof}

\hidden{
\begin{clemma}{15}
  \label{lem:min2}
  Let $t = C_1[C_2[C_3[C_4[t_5]]]] \in \bT$ and $t' = C_1[C_3[t_5]] \in \bT$ such that $t$ is $x, y, l$-separated and neither $C_2$ nor $C_4$ do generate string tokens in $\{x,y\}$. By Lemma~\ref{lem:downwardsep}, $t$ is separated. Let $C_x[C_0[t'_y]]$ be a minimal separation of $t'$. There exists a minimal separation $D_x[D_0[t_y]]$ of $t$ such that $D_x$ contains all nodes of $C_x$. 
\end{clemma}
\begin{proof}
  If $\height(C_0) = 0$, let $E_x[t'_1]$ be the split of $t'$ such that the root node of $t'_1$ is the highest node of $t'_y$. One checks easily that $E_x[X[t_1]]$ is a $0$-separation of $t$ (because $C_2$ and $C_4$ do not add $x$ or $y$-tokens), minimal because $\height(X) = 0$.
  
  If $\height(C_0) \neq 0$, assume for contradiction that a node $\pi \in C_x$ is not in $D_x$ after insertion of $C_2$ and $C_4$. $\pi$ must then lie in $D_0$, for otherwise the whole subtree of $t'$ rooted at $\pi$ would generate no $x$-tokens, thus so would the subtree of $t'$ rooted at $\pi$ and, by minimality, we would have $\height(C_0) = 0$. By lemma~\ref{lem:min1}, $\pi$ is not a node of $C_0$ so we can write $C_x = C'_x[C_\pi]$ with $\height(C_{\pi}) > 0$ and $C_{\pi}$ is rooted at $\pi$. We then have $D_x[D_0] = D'_{x}[D_{\pi}[D'_0]]$ where $D'_{x}[D_{\pi}]$ is constitued of $C'_x[C_\pi]$ plus some nodes of $C_2$ and/or $C_4$. Letting $E_x = D'_{x}[D_{\pi}]$, $E_x[D'_0[t_y]]$ is an $x, y, l$-separation of $t'$ which contradicts the minimality of $D_x[D_0[t_y]]$.
\end{proof}
}

\subsection{Inductive bounds}

For any tree or context $t$ and symbol $x$, let us write $n^t_x$ as a shorthand for $\countof{x}{\yield(t)}$, $e^t_x$ for the number of $x$-edges generated by $t$ and $r^t_x$ the length of the rightmost maximal substring of $\yield(t)$ consisting in only $x$-tokens (more formally, $r^t_x = \countof{x}{s}$, where $s$ is the unique substring such that $\yield(t) = u \cdot s \cdot v$ where $s \in x^*$, if $u$ is non empty its last token is not $x$, and $\countof{x}{v} = 0$).
  
  There is a maximal number of string tokens and edges that a context of height at most $p$ can generate under the considered yield and homomorphism. We call $l_0$ this number and focus from now on $l_0$-separated and $l_0$-asynchronous derivations.

Below are the proofs of the two statements of Lemma 7 of the main paper (respectively, 7-1 and 7-2).

\begin{clemma}{7-1}
  \label{lem:xbound}
  If $t \in \bT$ is $x,y,l_0$-separated and $t = D_x[D_0[t_y]]$ is an $x,y,l_0$-separation of $t$, then for $t_0 = D_0[t_y]$ we have
  \begin{equation}
    %&e^{t_0}_{\csdm{y}} \ge n^{t}_{\csdm{y}} - n^t_xl_0 - r^t_{\csdm{y}} \tag{$y$ bound}\\
    e^{t_0}_{\csdm{x}} \le n^{t_0}_{\csdm{x}} + n^{t}_xl_0 \tag{$x$ bound}.
  \end{equation}
\end{clemma}
\begin{proof}
  We prove the result by induction over the pair $(\height(t_0), \height(t))$ (with lexicographic ordering).

  \textbf{Base Case}
  Assume $\height(t_0) \le p$. Then $e^{t_0} \le l_0$. Since $\yield(t) \in \csd_s$, $n^t_x > 0$,  thus $n^{t_0}_{\csdm{x}} + n^t_xl_0 \ge l_0$ which ensures the bound.

  \textbf{Induction step}
  If $h(t_0) > p$ then $h(t) \ge h(t_0) > p$. We apply Lemma~1 to $t$ to yield a decomposition
  $t = C_1[C_2[C_3[C_4[t_5]]]]$, where $t' = C_1[C_3[t_5]] \in \bT$,
  $\height(t') < \height(t)$ and $\height(C_2[C_3[C_4]]]) \leq p$. Notice that $t_0$ cannot overlap with $C_2[C_3[C_4]$ without overlapping with $C_1$ or $t_5$ as well, for otherwise $h(t_0) \le p$.

  As in the proof of Lemma~\ref{lem:downwardsep}, letting $C_x, C_0, t'_y$ be obtained by removing all nodes from $C_2$ and $C_4$ from $D_x$, $D_0$ and $t_y$ respectively, we obtain an $x,y,l$-separation $t' = C_x[C_0[t'_y]]$. We let $t'_0 = C_0[t'_y]$ and distinguish between possible configurations for $C_2$ and $C_4$:

  \textbf{Case 0}
  If neither $C_2$ or $C_4$ generate any $\csdm{x}$-token, we find by induction \[e^{t'_0}_{\csdm{x}} \le n^{t'_0}_{\csdm{x}} + n^{t'}_xl_0.\] Moreover, we have $e^{t'_0}_{\csdm{x}} = e^{t_0}_{\csdm{x}}$, $n^{t'_0}_{\csdm{x}} = n^{t_0}_{\csdm{x}}$ and $n^{t'}_x \le n^{t}_x$ which concludes.
  
  \textbf{Case 1}
  In this case Lemma~\ref{lem:pluscase} applies \emph{i.e.} $C_2$ and $C_4$ generate only some $\csdm{z}$-tokens and brackets. The only subcase not already covered by Case 0 is the one where $\csdm{z} = \csdm{x}$. Notice that $n^{t'}_x = n^{t}_x$. By induction, \[e^{t'_0}_{\csdm{x}} \le n^{t'_0}_{\csdm{x}} + n^{t'}_xl_0.\] If $t_0$ does not overlap with $C_2$ or $C_4$, we have $e^{t'_0}_{\csdm{x}} = e^{t_0}_{\csdm{x}}$ and $n^{t'_0}_{\csdm{x}} = n^{t_0}_{\csdm{x}}$ which ensures the bound. Otherwise $t_0$ overlaps with $C_4$. If all nodes of $t_0$ are contained in $C_4[t_5]$, then by Lemma~\ref{lem:pluscase}, $t_0$ generate no $y$-token. By separation, neither does $t$ which contradicts $t \in \csd$. Hence $t_0$ contains all nodes of $C_4$. Then by lemma~\ref{lem:pluscase} again, $n^{C_2[C_4]}_{\csdm{x}} = n^{C_4}_{\csdm{x}}$, hence $n^{t_0}_{\csdm{x}} = n^{t'_0}_{\csdm{x}} + n^{C_2[C_4]}_{\csdm{x}}$ and $e^{t_0}_{\csdm{x}} \le e^{t'_0}_{\csdm{x}} + n^{C_2[C_4]}_{\csdm{x}}$ which yields \[e^{t_0}_{\csdm{x}} \le e^{t'_0}_{\csdm{x}} + n^{C_2[C_4]}_{\csdm{x}}\le n^{t_0}_{\csdm{x}} + n^{t}_xl_0.\]

  \textbf{Case 2}
  In this case Lemma~\ref{lem:corecase} applies and at least one of  $C_2$-$C_4$ generate some token $z \in \{a,b,c,d\}$. The only subcase not already dealt with in Case 0 is the one where we can set $z = x$. We thus get inductively: \[e^{t'_0}_{\csdm{x}} \le n^{t'_0}_{\csdm{x}} + n^{t'}_xl_0.\] Since $C_2$ or $C_4$ generate at least some $x$-token, we have $n^t_x  \ge n^{t'}_x+1$. Moreover $e^{t_0}_{\csdm{x}} \le e^{t'_0}_{\csdm{x}} + l_0$ since $C_2[C_4]$ generate at most $l_0$ $\csdm{x}$-edges, and $n^{t_0}_{\csdm{x}} \ge n^{t'_0}_{\csdm{x}}$. So we have $e^{t_0}_{\csdm{x}} \le n^{t'_0}_{\csdm{x}} + n^{t'}_xl_0 + l_0 \le n^{t_0}_{\csdm{x}} + n^{t}_xl_0$ concluding the proof.  
\end{proof}

\begin{clemma}{7-2}
  \label{lem:ybound}
  If $t \in \bT$ is $x,y, l_0$-separated then $t$ there exists a minimal $x,y, l_0$-separation $D_x[D_0[t_y]]$ of $t$ is such that, letting $t_0 = D_0[t_y]$, we have
  \begin{equation}
  e^{t_0}_{\csdm{y}} \ge n^{t}_{\csdm{y}} - n^t_xl_0 - r^t_{\csdm{y}} \tag{$y$ bound}
    %&e^{t_0}_{\csdm{x}} \le n^{t_0}_{\csdm{x}} + n^{t}_xl_0 \tag{$x$ bound}.
  \end{equation}
\end{clemma}
\begin{proof}
  We prove the result by induction over the height of $t$.  

  $t$ is $x,y, l_0$-separated so let us consider $D_x[D_0[t_y]]$ a minimal $x,y, l_0$-separation of $t$. Let $t_0 = D_0[t_y]$.
  
  \textbf{Base Case}
  Assume $\height(t) \le p$. Then. $n^t_{\csdm{y}} \le l_0$. Since $\yield(t) \in \csd_s$, $n^t_x > 0$. Moreover, $0 \le e^{t_0}_{\csdm{y}}$ and $n^{t}_{\csdm{y}} \le l_0$. So $n^{t}_{\csdm{y}} - n^t_{x}l_0 - r^t_{\csdm{y}} \le 0 \le e^{t_0}_{\bar y}$ which ensures the bound.
  
  \textbf{Induction step}
  If $h(t) > p$, we apply Lemma~1 to $t$ to yield a decomposition
  $t = C_1[C_2[C_3[C_4[t_5]]]]$, where $t' = C_1[C_3[t_5]] \in \bT$,
  $\height(t') < \height(t)$ and $\height(C_2[C_3[C_4]]) \leq p$. By Lemma~\ref{lem:downwardsep}, $t'$ is $x,y, l_0$-separated. We let $t' = C_x[C_0[t'_y]]$ be a minimal separation of $t'$ verifying the bound and $t_0' = C_0[t'_y]$. In other words, we have:
  \begin{equation}
    e^{t'_0}_{\csdm{y}} \ge n^{t'}_{\csdm{y}} - n^{t'}_xl_0 - r^{t'}_{\csdm{y}} \label{eq:ybi}.
    %&e^{t'_0}_{\csdm{x}} \le n^{t'_0}_{\csdm{x}} + n^{t'}_xl_0 \label{eq:xbi}
  \end{equation}
  By Lemma~\ref{lem:min1}, $t_0 = D_0[t_y]$ contains all nodes of $t'_0$. We distinguish cases according to Lemmas~\ref{lem:pluscase}~and~\ref{lem:corecase}.
  
  \textbf{Case 1}
  Consider first the case where Lemma~\ref{lem:pluscase} applies \emph{i.e.} $C_2$ and $C_4$ generate only one kind of bar token, $\csdm{z}$, and brackets. We now distinguish cases depending on the value of $\csdm{z}$. Before this, we emphasize that in all subcases it holds that $n^{t}_{x} = n^{t'}_x$.

  \textbf{subcase i) $\csdm{z} \neq \csdm{y}$.}
  Since all nodes of $t'_0$ are contained in $t_0$, we have $e^{t_0}_{\csdm{x}} \ge e^{t'_0}_{\csdm{x}}$.  Since $C_2$ and $C_4$ generate no $\csdm{y}$-token, we have $n^{t}_{\csdm{y}} = n^{t'}_{\csdm{y}}$ and $r^t_{\csdm{y}} = r^{t'}_{\csdm{y}}$. Injecting into inequation~\eqref{eq:ybi} concludes. 
  
  \textbf{subcase ii) $\csdm{z} = \csdm{y}$.}
  We distinguish the different possible overlap of $C_2$ and $C_4$ with $t_0$.
  Notice first that, by minimality, if any $C_i$, $i \in \{2,4\}$ overlaps with $t_0$ then $t_0$ contains all nodes of $C_i$, for otherwise we would have $D_0 = D_0'[D_0'']$ with $D_0'$ a subcontext of $C_i$ such that $\height(D_0' > 0)$, and in that case $(D_x[D_0'], D_0'', t_y)$ would be a smaller $x,y,l$-separation of $t$ since $C_i$ (hence $D_0'$) does not generate $y$-tokens.
  
  Hence, in the case where $t_0$ overlaps with $C_2$, $t_0$ contains all nodes of $C_2$ and $C_4$. Since $t_0$ also contains all nodes of $t'_0$, $e^{t_0} \ge e^{t'_0}+e^{C_2[C_4]} = e^{t'_0} + n^{t}_{\csdm{y}} - n^{t'}_{\csdm{y}}$. Moreover, $r^{t}_{\csdm{y}} \ge r^{t'}_{\csdm{y}}$. We can then conclude using inequation~\ref{eq:ybi}.

  Consider now the case where $t_0$ does not overlap with $C_2$ or $C_4$. Since all $y$-tokens are generated by $t_0$, projectivity of the yield and the definition of $\csd$ impose that $r^t_{\csdm{y}} = r^{t'}_{\csdm{y}} + n^{C_2[C_4]}_{\csdm{y}}$. We further have $e^{t_0}_{\csdm{y}} \ge e^{t'_0}_{\csdm{y}}$, and injecting into inequation~\ref{eq:ybi} yields $e^{t_0}_{\csdm{y}} \ge  n^t_{\csdm{y}} - n^{C_2[C_4]}_{\csdm{y}} + n^t_xl_0 -  r^{t_0}_{\csdm{y}} + n^{[C_2[C_4]}_{\csdm{y}}$ which simplifies into the desired $y$ bound.

    Finally, in the case where $C_2$ does not overlap with $t_0$ but $C_4$ does, all nodes of $C_2$ are contained in $D_x$ and all nodes of $C_4$ are contained in $t_0$. We must then have $\countof{y}{\yield(C_3)} > 0$. Otherwise, there would exist an $x,y,l$-separation $E_x[E_0[t_y]]$ with $E_x = C_1[C_2[C_3[C_4]]]$, and $\height(E_0) < \height(D_0)$.  Assume $\countof{x}{\yield(C_3)} > 0$. Lemma~\ref{lem:pluscase}, point 2, ensures that $\countof{x,y}{\leftc(C_3)} = 0$ or $\countof{x,y}{\rightc{(C_3)}} = 0$. Assume $\countof{x,y}{\rightc(C_3)} = 0$ (the other case is symmetric). We then have both a $x$ and a $y$ generated on the left of $C_3$. Since neither $C_1[C_2]$ nor $t_5$ generate any $y$-token, projectivity imposes $r^t_{\csdm{y}} = r^{t'}_{\csdm{y}} + n^{C_2[C_4]}_{\csdm{y}}$ and we can conclude as in the previous case. The only remaining subcase is when $\countof{x}{\yield(C_3)} = 0$, in which case $t$ is $0$-separated, and considering the (minimal) $0$-separation $(C_1, X, C_2[C_3[C_4]])$ we can use the same argument as in the case where $t_0$ encompasses all nodes of $C_2$ and $C_4$.

    \textbf{Case 2}
     Consider now the remaining case where Lemma~\ref{lem:corecase} applies. If neither $C_2$ or $C_4$ generate some $x$ or $y$-token, they don't generate $\csdm{x}$ or $\csdm{y}$-tokens either, and the same reasoning as Case 1 subcase i) applies. Otherwise $C_2[C_4]$ generate at least some $x$-token. We then have $n^{t}_x \ge n^{t'}_x + 1$. Since $t_0$ contains all nodes from $t'_0$ we further have $e^{t_0}_{\csdm{y}} \ge e^{t'_0}_{\csdm{y}}$. Finally $n^{t}_{\csdm{y}} \le n^{t'}_{\csdm{y}} + l_0$. We conclude using inequation~\ref{eq:ybi}.
\end{proof}

\subsection{Conclusion}

\begin{clemma}{8}
  For any $t \in \bT$, if $t$ is $x,y, l_0$-separated then $t$ is $x,y,l_0$-asynchronous.
\end{clemma}
\begin{proof}
  By Lemma~\ref{lem:ybound}, there is a minimal $x,y,l_0$-separation $t=D_x[D_0[t_y]]$ such that the $y$ bound obtains for $t_0 = D_0[t_y]$. By lemma~\ref{lem:xbound} the $x$ bound obtains for $t_0$ as well. Observe finally, that $r^{t}_{\csdm{y}} \le m^t_{\csdm{y}}$ and since $t_0$ generates at most $l_0$ $x$-tokens, by projectivity and definition of $\csd$, it generates at most $(l_0+1)m^t_{\csdm{x}}$ $\csdm{x}$-tokens (one sequence of $m^t_{\csdm{x}}$ between each occurrence of $x$ and the next, plus possibly one in front of the first and one after the last). Hence,
  \begin{align*}
    &e^{t_0}_{\csdm{y}} \ge n^{t}_{\csdm{y}} - n^{t}_xl_0- m^t_{\csdm{y}}\\
    &e^{t_0}_{\csdm{x}} \le  n^{t}_xl + m^t_{\csdm{x}}(l_0+1).
  \end{align*}
  and $t$ is $x,y,l_0$-asynchronous.
\end{proof}

\begin{clemma}{9}
  \label{lem:sep}
  For any $t \in \bT$, $t$ is $x,y, l_0$-separated for some $x/y \in \seps$.
\end{clemma}
\begin{proof}
  The proof proceeds by induction on the height of $t$.

  If $\height(t) \le p$. Then $\countof{z}{\yield(t)} \le l_0$ for any $z \in \{a,b,c,d\}$, hence $t$ is trivially $x,y,l_0$-separated for some $x/y \in \seps$.
  
  If $h(t) > p$, Lemma~1 yields a decomposition
  $t = C_1[C_2[C_3[C_4[t_5]]]]$, where $t' = C_1[C_3[t_5]] \in \bT$,
  $\height(t') < \height(t)$ and $\height(C_2[C_4]) \leq p$. By induction $t'$ is $x,y,l_0$-separated for some $x/y \in \seps$. For sake of succintness, let us present the inductive step for $x/y = a/c$, the reasoning for other cases is analoguous. Let us examine the different possible configurations of $C_2$ and $C_4$.

  \textbf{Case 1}
  If Lemma~\ref{lem:pluscase} applies \emph{i.e.} $C_2$ and $C_4$ generate only one kind of bar token, $\csdm{z}$, and brackets, one checks easily that inserting $C_2$ and $C_4$ does not change the distribution of $a$ and $c$-tokens in the tree, hence $t$ is $a, c, l_0$-separated.

  \textbf{Case 2}
  If Lemma~\ref{lem:corecase} applies, note first that if $C_2$ and $C_4$ generate no $a$ or $c$-token, we can conclude as in Case 1 as the distribution of $a$ and $c$-tokens in the tree is not changed either. Otherwise, we assume that $C_2$ or $C_4$ generate some $a$ or $c$-token and distinguish between subcases 1-3 of Lemma~\ref{lem:corecase}: 

  \textbf{Subcase 1} in this case for some $i \in \{2,4\}$ $\leftc(C_i)$ contains an $a$-token and no $b,c$ or $d$-token while $\rightc(C_i)$ contains some $c$-token and no $a,b$ or $d$-token. Assume $i=2$, the case where $i=4$ is similar. By projectivity and definition of $\csd_s$ follows that all $b$-tokens are generated in $C_3[C_4[t_5]]$ and all $c$-tokens in $C_1$. $t$ is therefore $b,d,0$-separated, hence $b, d, l_0$-separated.
  
  \textbf{Subcase 2} in this case, $\leftc(C_2)$ contains some $a$-token and no $b, c, d$-token, $\rightc(C_3)$ contains some $d$-token and no $a, c, d$-token, $\leftc(C_4)$ contains some $b$-token and no $a, c, d$-token, $\rightc(C_4)$ contains some $c$-token and no $a, b, d$-token. It follows that $t_5$ generate no occurrence of $a$ and $C_1$ no occurrence of $c$. Since $\countof{a}{\yield(C_2[C_3[C_4]])} \le l_0$, $(C_1, C_2[C_3[C_4]], t_5)$ is an $a,c,l_0$-separation.

  \textbf{Subcase 3} Assume $\leftc(C_2)$ contains some $a$-token and no $b, c, d$-token and that $\leftc(C_4)$ contains some $c$-token. It follows that all $b$-tokens are generated by $C_3$. So $\yield(t)$ contains less than $l_0$ $b$-tokens, by definition of $\csd$ it also contains less than $l_0$ $d$-tokens, so $(C_1, C_2[C_3[C_4]], t_5)$ is a $d, b, l_0$-separation.

  Assume now $\leftc(C_2)$ contains some $a$-token and no $b, c, d$-token and that $\rightc(C_4)$ contains some $c$-token. It follows that $t_5$ generate no $d$-token and $C_1$ generate no $b$-token. Hence $(C_1, C_2[C_3[C_4]], t_5)$ is $b,d, l_0$-separation.

  The remaining cases are symmetric exchanging $c$ with $a$, $d$ with $b$, and ${\leftc}$ with ${\rightc}$ everywhere.  
\end{proof}

\hidden{
\todo{*****************************OLD PROOF*******************************}

\hidden{

The proof below includes cases left out in the main discussion. To cover these cases one needs two additional lemmas that were not discussed in the main paper.
\begin{clemma}{8}
  Let $l_0$ bound the number of string tokens generated by a context
  of height at most $p$, where $p$ is the pumping height of $\bT$.
  %For any $t \in \bT$, if $t = C[t']$ such that $\countof{x}{\yield(t')} \le n$ and $\countof{\csdm{x}}{\yield(t)} \le r$, then $t = C[C'[t']]$ where $\height(C'') < t$ and $C''[t']$ generate at most $(n+1)l_0 + r$ $\csdm{x}$-edges.
  %It follows that
  For any $t \in \bT$, if $t = C[t']$ such that $\countof{x}{\yield(t)} \le n$ and $\countof{\csdm{x}}{\yield(t')} \le r$, $t$ generate at most $nl_0 + r$ $\csdm{x}$-edges.
  \label{lem:abound}
\end{clemma}

This is a consequence of the pumping lemma and the design of $\csd$: if $t$ is such that $|\yield(t)| > l_0$ the pumping lemma (Lemma~1) applies to $t$. This identifies two contexts $C_2$ and $C_4$ of $t$ which can be removed to produce a smaller tree $t'$, or be `iterated' in the sense described in Lemma~1. Since $t,t' \in \bT$ (still from Lemma~1), we can observe that $C_2$ and $C_4$ must generate all edges which are in $\graphof(t)$ but not $\graphof(t')$, and from the structure of $\csd$, that $C_2$ and $C_4$ must generate exactly as many $z$-edges that they produce $z$-tokens, for any symbol $z$.

Because $C_2$ and $C_4$ can be `iterated' in the sense of Lemma~1,
there are only two possibilities for $C_2$ and $C_4$ to contain
occurrences of $\csdm{x}$-symbols. If either one of $C_2$ or $C_4$
contains some occurrences of letters outside
$\{ \csdl, \csdm{x}, \csdr \}$, for example, $\leftc(C_2)$ contains an
occurrence of $a$, then $\leftc(C_2)$ can only contain some entire
well-bracketed words from $\csdl^k \csdm{x}^k \csdr^{k}$, otherwise
words outside $\csd_s$ would be produced by iteration of $C_2$ and
$C_4$. Conversely, if, say, $\leftc(C_2)$ contains some entire
well-bracketed words from $\csdl^k \csdm{x}^k \csdr^{k}$ this word
must occur on the left on an $x$ symbol in $\leftc(C_2)$ (otherwise
iterating $C_2$ and $C_4$ would yields words outside of
$\csd_s$). Moreover, the total number of symbols generated by $C_2$
and $C_4$ is always less than $l_0$. The other possibility is that
$C_2$ and $C_4$ both contains only symbols from
$\{ \csdl, \csdm{x}, \csdr \}$, but then both have to be nonempty and
occur on different sides in $C_2$ and $C_4$ (for instance, $C_2$ can
generate $\csdl\ \csdm{x}$ on the left but then $C_4$ has to generate
$\csdr$ on the right), and from projectivity constraints this means
that the whole subtree starting at $C_2$ contains only symbols in
$\{ \csdl, \csdm{x}, \csdr \}$. When considering sequentially removing
$\csdm{x}$-edges through the pumping lemma, instances of the first
cases above might yield alignments from $\csdm{x}$ tokens to symbols
in $\{a,b,c,d\}$, but this kind of removing might only happen a total
of $n$ times. The second kind produces alignments of $\csdm{x}$-edges
to $\csdm{x}$ tokens. Hence a subtree $t'$ generating a word such that
$\countof{\csdm{x}}{\yield(t')} \le r$ might generate, for each
occurrence of $\{a,b,c,d\}$ in $\yield(t)$, up to $l_0$
$\csdm{x}$-edges that were aligned to non $\csdm{x}$ string
tokens. And it can generate up to $r$ additional $\csdm{x}$-edges if
all its $\csdm{x}$ tokens are aligned with such edges.
}

The following lemma is useful to deal with a lot of cases in the proof of Lemma~7.

\begin{clemma}{9}
  If $t \in \bT$ and $t = C_x[C_0[t_y]]$ is a minimal
  $x,y,l$-separation and an $x,y,l$-asynchronous split, $C_1$ and
  $C_2$ generate no string tokens in $\{x,y,\bar x, \bar y\}$ and no
  edges labeled with these symbols, and $t'$  
  is obtained by inserting $C_1$ and $C_2$ into some positions of
  $t \in \bT$, then $t$ has a minimal separation
  $x, y, l$ separation and is also an $l$ asynchronous split.
\end{clemma}
\begin{proof}
  The definition of being separated is not impacted by tokens outside of $\{x,y,\bar x, \bar y\}$, so $t'$ is separated. Consider
  $D_x[D_0[t_y]]$ a $x,y,l$ minimal separation of $t'$. By a minimality argument, $D_0[t_y]$ contains all nodes from $C_0[t_y]$.
  Since the requirement of $x,y,l$ asynchronicity not affected by edge-labels outside of $\{x,y,\bar x, \bar y\}$, $D_0[t_y]$ validates these requirements.
\end{proof}

\begin{clemma}{7}
  If $\bT$ is an LM-CFTL, then there exists $l_0 \in \nats$ such that
  for any $t \in \bT$:
  \begin{enumerate}
  \item $t$ is $x,y,l_0$-separated for some $x,y \in \seps$.  
  \item For any $x,y \in \seps$, if $t$ is $x,y$, $l_0$-separated,
    there is a minimal $l_0$-separation $(C_x, C_0, t_y)$ of
    $t$ such that $(C_x, C_0[t_y])$ is an
    $x,y,l_0$-asynchronous split of $t$.   
  \end{enumerate}
  \label{lem:induction}
\end{clemma}
\begin{proof}
  Let $p$ be the pumping height of $\bT$, and let $l_0$ bound from
  above both the maximal number of string tokens and edges generated
  by a context of height $p$. We prove 1 and 2 by induction over
  $\height(t)$.

  \textbf{Base case.} If $\height(t) \le p$, $t$ is trivially
  $x,y,l_0$-separated for some $x/y \in \seps$ (point 1). 
  For point 2, let $t = C_x[C_0[t_y]]$ be a minimal
  $x,y,l_0$-separation of $t$. For $t_0 = C_0[t_y]$, $\graphof(t_0)$
  has at most $l_0 \leq (n_x+m)l_0$ $\csdm{x}$-edges. The lower bound
  on $\csdm{y}$-edges is trivially satisfied because there are at most
  $l_0$ tokens and edges. Thus $(C_x,t_0)$ is an
  $x,y,l_0$-asynchronous split.

  \textbf{Induction step.} If $\height(t) > p$, we apply
  Lemma~1 to $t$ to yield a decomposition
  $t = C_1[C_2[C_3[C_4[t_5]]]]$, where $t' = C_1[C_3[t_5]] \in \bT$,
  $\height(t') < \height(t)$ and $\height(C_2[C_4]) \leq p$.  By
  induction, there exist $x,y \in \seps$ such that $t'$ is
  $x,y,l_0$-separated. By point 2, some minimal separation
  $t'= C_x[C_0[t_y]]$ is an $x,y,l_0$-asynchronous
  split.

  The challenge is now to conclude 1 and 2 for $t$, i.e.\ after
  reinserting $C_2$ and $C_4$ into $t'$. These contexts may overlap
  with $C_x$ and $C_0$ in many ways; we can use pumping considerations
  to reduce these to four cases which differ in the tokens of
  $\yield(t)$ that are generated by $C_2$ and $C_4$. These four
  cases are displayed in Fig.~\ref{fig:cases}. 

  \textbf{Case 1.} Consider the case where for at least one
  $i\in \{2,4\}$, $\yield(C_i)$ contains at least one $x$ and at least
  one $y$, and that one of them is in $\leftc(C_i)$ and the other in
  $\rightc(C_i)$. Let's say that $x=a$, $y=c$, and $i=2$; the other
  cases are analogous.
% and at most $l_0$   edges. 
%Considering
  %  $\graphof(t)$ and $\graphof(t')$ we conclude than they generate a
%  total of at most $l_0$ $\csdm{a}$- or $\csdm{c}$-edges.
  % -> ?
  By projectivity and the definition of $\csd_s$, all $b$ letters must
  be generated in $C_3$ and $t_5$, and all $d$ letters in $C_1$. Thus, $t$
  and $t'$ are both $d,b,0$-separated. This establishes Point 1.

  For point 2, assume that $t$ is $a,c,l_0$-separated; $c/a$ is
  analogous and $b/d$ and $d/b$ are covered by Lemma 9. Let $t' = C_a[C_0[t_c]]$
  and $t = D_a[D_0[d_c]]$ be minimal $a,c,l_0$-separations. Let us
  write $e^{t}_z$ for the number of $z$-edges in $\graphof(t)$. By the
  induction hypothesis, $t'$ is $a,c,l_0$-asynchronous, so we have
  $e^{t'_1}_{\csdm{c}} \geq n^{(t')}_{\csdm{c}} - n^{(t')}_c
  l_0$. Furthermore, $\yield(t)$ has at least one more $c$-token than
  $\yield(t')$ (from $C_2$) and at most $l_0$ more $\csdm{c}$-tokens
  (from the pumping height), so
  $ n^{(t')}_{\csdm{c}} - n^{(t')}_c l_0 \geq n^{(t)}_{\csdm{c}} -
  n^{(t)}_c l_0$. One can show from the minimality of the separation
  of $t'$ that all the nodes contained in $t'_1 = C_0[t_c]$ must also
  be contained in $t_1 = D_0[d_c]$; thus
  $e^{t_1}_{\csdm{c}} \geq e^{t'_1}_{\csdm{c}}$, establishing the
  $\csdm{c}$ condition for the $a,c,l_0$-asynchronous split
  $(D_a,t_1)$ of $t$. 

  \textbf{Case 2.} Now consider the case where $\yield(C_2)$ and
  $\yield(C_4)$ only contain $\csdl$, $\csdm{y}$, and $\csdr$
  tokens. Adding $C_2$ and $C_4$ to $t'$ does not change the
  distribution of $x$ and $y$ tokens over the tree, so $t$ is still
  $x,y,l_0$-separated (point 1).

  Now let $t$ be $x',y',l_0$-separated for some $x'/y' \in
  \seps$. Cases where $y \neq x'$ and $y \neq y'$ are covered by Lemma
  9.  We consider first $y=y'$.  Let
  $t' = C_{x'}[C_0[t_{y'}]]$ be a minimal separation. By a minimality
  argument, we can find a minimal ${x'},{y'},l_0$-separation
  $t = C_{x'}[D_0[t_{y'}]]$, i.e.\ with the same outer context. Thus,
  $t_1 = D_0[d_{y'}]$ is constructed from $t'_1 = C_0[t_{y'}]$ by
  adding $C_2$ and $C_4$ somewhere.

  By induction hypothesis, $C_{x'}[t'_1]$ is an ${x'},{y'},l_0$-asynchronous
  split of $t'$, i.e.\
  $e^{(t'_1)}_{\csdm{{y'}}} \geq n^{(t')}_{\csdm{{y'}}} - n^{(t')}_{x'} l_0$. Let
  $e = e^{(t)}_{\csdm{{y'}}} - e^{(t')}_{\csdm{{y'}}}$ be the number of
  $\csdm{{y'}}$-edges added by $C_2$ and $C_4$; we have
  $n^{(t)}_{\csdm{{y'}}} - n^{(t')}_{\csdm{{y'}}} =e$ because
  $t,t' \in \bT$. Thus we have
  $e^{(t_1)}_{\csdm{{y'}}} = e^{(t'_1)}_{\csdm{y'}} + e \geq
  n^{(t')}_{\csdm{y'}} + e - n^{(t')}_{x'} l_0 = n^{(t)}_{\csdm{y'}} -
  n^{(t)}_{x'} l_0$. Because adding $C_2$ and $C_4$ changes neither the
  token count nor the edge count for ${x'}$ and $\csdm{{x'}}$, it follows that
  $(D_a,t_1)$ is an ${x'},{y'},l_0$-asynchronous split for $t$.

  Consider now the case $y=x'$. Removing only $\csdm{x}'$ tokens from $t$ doesn affect separation.
  So $t'$ is also $x'/y'$ separated.  Let $t' = E_{x'}[E_0[t_{y'}]]$ be a minimal
  $l_0$-separation. By a minimality argument, we can find a minimal
  ${x'},{y'},l_0$-separation $t = D_{x'}[E_0[t_{y'}]]$.  The argument
  for the $\csdm{c}$-condition is then simialr as above.
  The $\csdm{a}$-condition has been ensured to follow from separation
  by Lemma 8 in the preamble of this discussion.

  \textbf{Case 3.} Now consider the case where $\yield(C_2)$ and
  $\yield(C_4)$ only contain $\csdl$, $\csdm{x}$, and $\csdr$
  tokens. Similarily to case 2, adding $C_2$ and $C_4$ to $t'$ does not change the
  distribution of $x$ and $y$ tokens over the tree, so $t$ is still
  $x,y,l_0$-separated (point 1).

  Now let $t$ be $x',y',l_0$-separated for some $x'/y' \in \seps$. Again, cases where $y \neq x'$ and $y \neq y'$ are covered by Lemma
  9.

  We consider first $x=x'$. Up to the renaming $x' \rightarrow y', y' \rightarrow x'$, this case is the same as case 2 where $y=x'$.
  Consider now $x = y'$. Up to the renaming $x' \rightarrow y', y' \rightarrow x'$, this case is the same as case 2 where $y=y'$.

  \textbf{Case 4.} In this case, $C_2$ generates at least one $x$ symbols and no $y$ symbols and $C_4$ generate at least one $y$ symbol and no $x$ symbols. This case follows the same line of reasoning as case 2 with tiny variations (when $t$ is not $a/c$ separated one can't conclude to $0, b,d$-separation but only to $l_0, b, d$ separation).

  \hidden{
  \textbf{Case 5.} This case occurs when $\yield(C_2)$ and $\yield(C_4)$ both contain only symbols in $\{ \csdl, \csdm{x}, \csdr \}$.  %both $C_2$ and $C_4$ 
%  are inserted in positions of $C_0$, \emph{i.e.} letting
%  $D_0 = C_0'[C_2[C_0''[C_4[C_0''']]]]$, $t=C_x[D_0[t_y]]$, where
%  $C_0'[C_0''[C_0''']] = C_0$. Since $C_2$ and $C_4$ generate neither $x$ nor $y$ symbols,
%  $\langle C_x, D_0, t_y \rangle$ must be a minimal $l_0$-separation of $t$.

  Observing $\graphof(t)$ and $\graphof(t')$, we conclude that $C_2$ and $C_4$ generate exactly as many $\csdm{x}$-edges than they generate $\csdm{x}$ symbols. Let $q$ be this number. From height constraints, $q \le l_0$. Let $n_x = \countof{x}{\yield(t)} = \countof{x}{\yield(t')}$ (neither $C_2$ nor $C_4$ generate any $a$s). Let $t_0 = C_0[t_y]$ and $t_1 = D_0[t_y]$. Let $s_0 = \yield(t_0)$, $s_1 = \yield(t_1)$, and let $m_0$ (resp. $m_1$) be the maximal length of a contiguous sequence of only $\csdm{x}$ symbols in $t_0$ (resp. $t_1$). From pumping considerations on $C_2$ and $C_4$, we see that insertion of $C_2$ and $C_4$ never `splits' a sequence of $\csdm{x}$ symbols into two sequences separated by a symbol which is not $\csdm{x}$, nor insert symbols into two different sequences of $\csdm{x}$ symbols separated by another symbol. We therefore have $m_{1} \ge m_{0}$.

  If $m_{0} = m_{1}$, this means that the insertion of $C_2$  and $C_4$ did not increase the maximal length of a sequence of $\csdm{x}$ symbols going from $s_{0}$ to $s_{1}$, and since $\countof{x}{s_{C_0}} \le l_0$ , there is at most $l_0m_{0} - q$ $\csdm{x}$ symbols in $s_{0}$. By induction hypothesis point 3, the number $e_{0}$ of $\csdm{x}$-edges generated by $t_0$ is then such that $e_{0} \le n^{(t_0)}l_0 + l_0m_{0} - q$. From this follows that the number $e_{1}$ of $\csdm{x}$-edges generated by $t_1$ is such that $e_{1} \le (n^{(t_0)}l_0 + l_0m_{C_0} \le (n^{(t)} + r^{(t)})l_0$ which validates the $\csdm{x}$-condition for asynchronicity.

  If $m_{0} \ge m_{1} + 1$, we have by induction hypothesis that $e_{0} \le n^{(t_0)}l_0 + l_0m_{0}$ and since $e_1 = e_0 + q \le e_0 + l_0$ we have $e_{1} \le n^{(t_0)}l_0 + l_0m_{0} + l_0 \le n^{(t_0)}l_0 + l_0m_{1} \le n^{(t)}l_0 + r^{(t)}l_0$.
  } 
\end{proof}

\subsection{Separation}

%%% Local Variables:
%%% mode: latex
%%% TeX-master: "supplementary"
%%% End:
}